\documentclass{article}

\PassOptionsToPackage{numbers, compress}{natbib}
\usepackage[preprint]{neurips_2024}
\usepackage{fontawesome}

\usepackage{longtable}

\usepackage[utf8]{inputenc} 
\usepackage[T1]{fontenc}    
\usepackage[colorlinks,
            linkcolor=red,
            anchorcolor=blue,
            citecolor=green
            ]{hyperref}
\usepackage{url}            
\usepackage{booktabs}       
\usepackage{amsfonts}       
\usepackage{nicefrac}       
\usepackage{microtype}      
\usepackage{xcolor}         
\usepackage{caption}
\usepackage{subcaption}
\usepackage{natbib}
\usepackage{graphicx}
\usepackage{enumitem}
\usepackage{makecell}
\usepackage{amsmath}
\usepackage{amsthm}
\usepackage{thmtools}
\usepackage{thm-restate}
\usepackage{algorithm}
\usepackage{algorithmic}
\usepackage{bbm}
\usepackage{tabularx}
\usepackage{listings}
\usepackage{xcolor}
\usepackage{multirow}
\usepackage{multicol}
\usepackage{tcolorbox}
\usepackage{tikz}
\usepackage{footnote}
\usepackage{wrapfig}
\usepackage{tikz}

\definecolor{tblue}{RGB}{31,119,180}
\definecolor{torange}{RGB}{255,127,14}
\definecolor{tgreen}{RGB}{44,160,44}
\definecolor{tred}{RGB}{214,39,40}
\definecolor{tpurple}{RGB}{148,103,189}
\definecolor{lightblue}{RGB}{173, 216, 230}
\definecolor{lightpink}{RGB}{255, 182, 193}
\definecolor{lightgreen}{RGB}{144, 238, 144}

\usepackage{colortbl}
\usepackage{xcolor}
\usepackage{array}
\usetikzlibrary{fadings}

\newcommand{\hide}[1]{} 

\newcommand{\eg}{\textit{e}.\textit{g}.} 
 
\newcommand{\ag}[1]{\texttt{\textbf{#1}}}

\def\model{AI-Researcher}
\def\bench{Scientist-Bench}

\title{AI-Researcher: Autonomous Scientific Innovation}
\lstdefinelanguage{Prompt}{
  moredelim=[s][\color{violet}\bfseries]{`}{`}, 
  moredelim=[s][\color{violet}\bfseries]{<}{>}, 
  morecomment=[s]{<!--}{-->}, 
  morestring=[b]", 
  stringstyle=\color{blue!60}, 
  commentstyle=\color{green}\itshape, 
}
\lstdefinelanguage{Tools}{
literate={[}{{\color{violet}\bfseries[}}1
           {]}{{\color{violet}\bfseries]}}1, 
  morecomment=[s]{<!--}{-->},  
  morestring=[b]",  
  stringstyle=\color{cyan!60},  
  commentstyle=\color{green}\itshape,  
}
\author{
  Jiabin Tang\thanks{Equal contribution.}\footnotemark[1] ~~~
  Lianghao Xia\footnotemark[1] ~~~
  Zhonghang Li~~~
  Chao Huang\thanks{Chao Huang is the Corresponding Author.} \\
  The University of Hong Kong \\
  \texttt{\{jiabintang77, bjdwh.zzh, chaohuang75\}@gmail.com; aka\_xia@foxmail.com}\\
  \faGithub~\textbf{Source Code:} \textcolor{blue}{\url{https://github.com/HKUDS/\model}}
}

\begin{document}

\maketitle

\begin{abstract}
The powerful reasoning capabilities of Large Language Models (LLMs) in mathematics and coding, combined with their ability to automate complex tasks through agentic frameworks, present unprecedented opportunities for accelerating scientific innovation. In this paper, we introduce \model, a fully autonomous research system that transforms how AI-driven scientific discovery is conducted and evaluated. Our framework seamlessly orchestrates the complete research pipeline--from literature review and hypothesis generation to algorithm implementation and publication-ready manuscript preparation--with minimal human intervention. To rigorously assess autonomous research capabilities, we develop Scientist-Bench, a comprehensive benchmark comprising state-of-the-art papers across diverse AI research domains, featuring both guided innovation and open-ended exploration tasks. Through extensive experiments, we demonstrate that \model\ achieves remarkable implementation success rates and produces research papers that approach human-level quality. This work establishes new foundations for autonomous scientific innovation that can complement human researchers by systematically exploring solution spaces beyond cognitive limitations.
\end{abstract}


\section{Introduction}\label{sec:intro}

Scientific discovery has historically been constrained by human cognitive limitations and the immense scale of potential solution spaces~\cite{wang2023scientific}. Recent advances in Large Language Models (LLMs) have demonstrated remarkable capabilities in mathematical reasoning, coding, and problem-solving tasks that were previously thought to require human expertise~\cite{didolkar2024metacognitive,guo2024deepseek}. However, transitioning from isolated capabilities to fully autonomous scientific research systems capable of original innovation remains an unsolved challenge that could fundamentally transform how scientific progress occurs.

Despite recent advances in agentic frameworks powered by LLMs, scientific innovation represents an intellectual frontier orders of magnitude more challenging than the task automation currently mastered by existing AI agents~\cite{manus2025, openmanus2025, li2023CAMEL, tang2025AutoAgent}. While today's agents can schedule meetings or retrieve structured information, genuine scientific discovery demands an unprecedented level of intelligence—requiring sophisticated conceptual reasoning across abstract theoretical domains, transformative hypothesis generation that bridges disparate knowledge fields, and methodological innovation that extends far beyond pattern recognition. The research process necessitates maintaining coherent understanding across thousands of papers while simultaneously generating insights that fundamentally advance knowledge boundaries—intellectual capabilities that existing architectures cannot approach.

Most critically, scientific exploration involves navigating vast, unbounded solution spaces with deeply uncertain rewards, requiring meta-cognitive abilities to recognize promising directions and abandon unproductive paths. Researchers must continuously evaluate experimental results against theoretical frameworks, adapt hypotheses based on unexpected findings, and communicate complex ideas with precision and clarity—all while maintaining the creative spark that drives breakthrough discoveries. These profound limitations have prevented AI systems from autonomously conducting meaningful scientific work, perpetuating a paradigm where AI remains relegated to narrow assistance roles rather than functioning as independent scientific contributors capable of accelerating human knowledge advancement through systematic exploration of solution spaces beyond human cognitive limitations.

\begin{figure*}[h]
    \centering
    \includegraphics[width=0.97\linewidth]{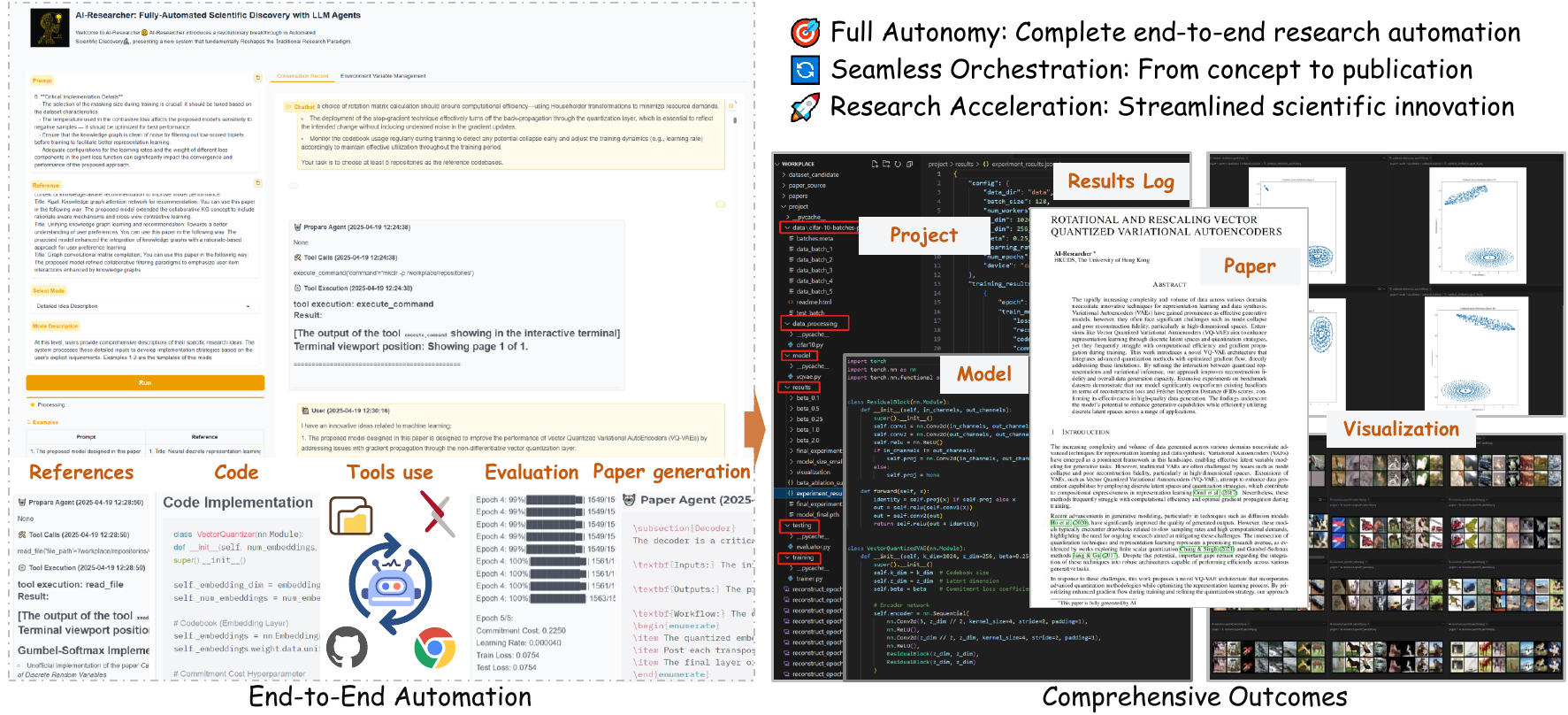}
    \caption{Architectural overview of AI-Researcher, illustrating the end-to-end autonomous scientific innovation pipeline encompassing literature exploration, idea generation, algorithm implementation, experimental validation, and comprehensive scholarly publication with rigorous evaluation metrics.}
    \label{fig:intro}
\end{figure*}

While specialized systems exist for literature analysis or experiment design~\cite{schmidgall2025AgentRxiv,gottweis2025towards}, they fail to orchestrate the complete research workflow from hypothesis generation through publication-quality reporting. Furthermore, no standardized benchmarks exist to evaluate autonomous research across diverse scientific domains, making progress in this frontier difficult to measure systematically.

We introduce \model\, a novel framework that addresses these limitations by seamlessly orchestrating the complete scientific discovery lifecycle—from literature analysis through implementation to scholarly documentation. Unlike systems focusing on isolated capabilities, our framework employs a comprehensive multi-agent architecture where specialized components collaborate through structured knowledge exchange to maintain coherent reasoning throughout the research process. This recursive refinement mechanism enables continuous bidirectional feedback between theoretical concepts and their implementations—preserving intellectual consistency while transforming research ideas into rigorous scientific contributions with minimal human intervention.

\model\ introduces three key innovations that fundamentally advance autonomous scientific discovery. \textbf{First}, Resource Analyst agents decompose complex research concepts into atomic components with explicit bidirectional mappings between mathematical formulations and code implementations, dramatically reducing hallucination risks. \textbf{Second}, our Implementation Framework employs a human-inspired iterative refinement paradigm where specialized agents collaborate through structured feedback cycles, mirroring the proven mentor-student relationship in academic research. \textbf{Third}, our Documentation Agent overcomes LLM coherence limitations through a hierarchical synthesis approach that transforms research artifacts into publication-quality manuscripts while maintaining cross-document consistency and factual integrity throughout extensive scholarly documentation.

To rigorously evaluate autonomous scientific systems, we develop \textbf{Scientist-Bench}—the first comprehensive benchmark enabling standardized assessment across both guided innovation scenarios and open-ended exploration tasks spanning diverse AI domains. Through extensive experiments on 22 benchmark papers using multiple LLM evaluators, we demonstrate that \model\ achieves remarkable implementation success rates while producing research contributions that frequently approach human-level quality. Surprisingly, our findings reveal \model\ performs better in open-ended exploration than in guided implementation tasks—suggesting autonomous research systems excel when leveraging internal knowledge synthesis rather than following prescriptive directives. These results establish new foundations for autonomous scientific agents that complement human researchers by systematically exploring solution spaces beyond human cognitive limitations.




\section{Scientist-Bench: Benchmarking AI Agents for Scientific Discovery}
\label{sec:bench}

Scientific discovery represents one of humanity's most complex intellectual endeavors, requiring creativity, methodical reasoning, and expertise. {Developing benchmarks for novel scientific discovery and establishing multifaceted evaluation metrics remain critical methodological challenges in the field~\cite{reddy2025towards}.} To advance AI-assisted research, we introduce \bench, a comprehensive benchmark enabling direct comparison between LLM Agent-generated research outcomes and high-quality scientific work produced through human expertise. In contrast to other related benchmarks~\cite{wang2024scibench}, our benchmark provides carefully curated research instructions, relevant references, contextual datasets, and ground-truth research papers for multidimensional evaluation across various scientific criteria.

By leveraging peer-reviewed papers from AI/Data Science top-tier conferences as reference standards and extracting their core research questions as agent instructions, \bench\ creates a robust framework to systematically measure how AI-generated scientific contributions compare to human-produced insights across multiple dimensions of research quality. This approach enables a meaningful assessment of how close AI systems can come to replicating or complementing human scientific achievement. We begin by formally defining the scientific discovery task as follows:

\subsection{Task Formulation}

\noindent\textbf{Agent System Input}. For each sample in \bench, we use a target paper $y$ authored by human researchers as the evaluation standard. The input features $\mathcal{X}=\{\mathcal{R}, I, \mathcal{D}\}$ comprise reference papers $\mathcal{R}$ (15-20 relevant references from paper $y$ selected via LLMs), a research instruction $I$ (containing the core research idea extracted from $y$), and datasets $\mathcal{D}$. To evaluate innovation capabilities, \bench\ defines two distinct challenge levels: \textbf{Level-1} tasks provide explicit research instructions directly extracted from paper $y$, testing agents' ability to execute given ideas; \textbf{Level-2} tasks deliberately omit these instructions, challenging agents to independently formulate novel research directions using only the provided references and datasets. Our benchmark samples span diverse research fields including diffusion models, vector quantization, graph neural networks, and recommendation systems. Prompts used to construct this input data are detailed in Appendix~\ref{sec:bench_cons}.

\noindent\textbf{Agent System Output}. The output $\hat{\mathcal{Y}}=\{\mathcal{C}, p\}$ consists of code scripts $\mathcal{C}$ that implement the research proposals and a technical report $p$ describing the research background, motivation, methodology, experiments, and results. Both components undergo assessment through \bench's evaluation module to measure the quality and innovation of the agent's scientific contributions compared to human-generated research work. This dual evaluation of implementation and documentation provides a holistic view of the agent's capabilities across both theoretical and practical dimensions.




\subsection{Benchmark Construction}

\noindent $\bullet$ \textbf{Step 1: Systematic Selection of AI Research Benchmark Papers}. 
To establish a comprehensive evaluation framework for AI research systems, we systematically collected papers from 2022-2024 spanning expertise levels across diverse domains. Our methodology employed a two-pronged approach for identifying high-impact contributions: \textbf{First}, we leveraged LLMs to generate domain-specific keywords across 16 research areas including ``Computer Vision'', ``Graph Learning'', ``Recommender Systems'', ``Vector Quantization'', ``Image Processing'', ``Self-Supervised Learning'', ``Contrastive Learning'', and others.  \textbf{Second}, we retrieved top-cited papers from arXiv for each domain (10 papers per keyword) and applied citation-based filtering metrics. This process culminated in selecting 22 representative papers that showcase breakthrough research across the AI landscape, providing a robust foundation for evaluating AI systems on scientific discovery and research comprehension.

\noindent $\bullet$ \textbf{Step 2: Input Construction for AI-Researcher}.
To generate input for AI Agent systems, we emulate the scholarly research approach of first reviewing literature extensively before formulating new directions. We construct input information from two complementary dimensions: i) \textbf{Reference Literature Review}: domain-specific references providing knowledge foundation; and ii) \textbf{Research Requirements}: strategic objectives that direct the Agent toward targeted scientific discovery paths.

\noindent \textbf{Reference Literature Review}. Understanding the scientific research process is essential for developing effective AI research systems. Just as human researchers begin by exploring relevant literature before conducting their own investigations, our AI-Researcher model follow a similar path. Identifying key influences on scientific advancement requires rigorous methodology. Our process aims to distill the 15-20 references $\mathcal{R}$ that fundamentally influenced each target paper $y$, revealing the intellectual foundations upon which breakthroughs are built. By extracting these critical references, we construct appropriate inputs for our AI-Researcher that mirror the human research process.

We prioritize references that provide methodological frameworks, contribute essential components, or inspire conceptual innovations—elements that illuminate the paper's scientific lineage. To ensure both systematic rigor and objective assessment of reference importance, we have implemented a comprehensive five-step LLM-based evaluation process for reference input selection as follows: i) \textbf{Citation pattern analysis}: Quantify citation frequency and section distribution to identify strategically placed references.
ii) \textbf{Context analysis}: Evaluate each reference's influence on methodology, theory, and experimental design.
iii) \textbf{Evidence collection}: Gather specific textual evidence demonstrating reference impact for transparent verification.
iv) \textbf{Impact scoring}: Compute importance scores through integrated analysis of quantitative and qualitative factors.
v) \textbf{Final selection}: Select and justify the top 15-20 references that demonstrably shaped the paper's contributions.


\noindent\textbf{Research Requirement Generation}. To formulate the research directive $I$, we employ LLMs to extract the fundamental research concept from the target paper $y$. This systematic extraction identifies the core research focus, existing limitations, critical challenges, and primary objectives--effectively capturing the study's essential contributions and underlying motivations. To maintain scientific integrity and prevent information leakage, we carefully exclude all technical specifications, model identifiers, quantitative results, and architectural details from the research directive.


\noindent $\bullet$ \textbf{Step 3: Rigorous Anonymization to Ensure Scientific Originality}. A critical challenge in evaluating AI research agents lies in distinguishing between genuine problem-solving abilities and mere regurgitation of memorized content. To address this fundamental concern, we implement a comprehensive anonymization protocol that transforms the evaluation landscape: i) \textbf{Method name masking}: Replace algorithm and model names with generic identifiers, testing conceptual understanding rather than term recognition. ii) \textbf{Technical detail abstraction}: Remove implementation specifics while preserving core concepts, requiring engagement with fundamental principles. iii) \textbf{Dataset standardization}: Normalize experimental contexts to create a fair evaluation landscape that prevents shortcuts based on dataset familiarity. iv) \textbf{Citation anonymization}: Eliminate temporal and institutional markers to test problem-solving rather than information recall.

\subsection{Evaluation of AI-conducted Scientific Discovery}
To rigorously assess the genuine scientific discovery capabilities of AI agent system on our \bench\ benchmark, we implement a two-stage evaluation framework that addresses both technical implementation fidelity and scientific innovation merit.

\noindent $\bullet$ \textbf{Stage 1: Technical Execution Validation}. 
The first stage employs a specialized code review agent to verify whether the implementation code $\mathcal{C}$ faithfully realizes the AI-conducted research innovations. This critical verification prevents scenarios where AI researchers might propose sophisticated methods with promising results without providing functional implementations—a fundamental requirement for credible scientific discovery. The code review agent performs static analysis and runtime verification across key dimensions including \emph{Algorithm Correctness}, \emph{Computational Efficiency}, and \emph{Adherence to Specified Constraints}. We quantify this assessment using a completion ratio metric that reflects the proportion of required functionality successfully implemented by the AI researcher.

\noindent $\bullet$ \textbf{Stage 2: Scientific Contribution Evaluation.}
The second stage rigorously assesses whether AI agent systems have produced genuine scientific innovations by comparing the generated research report $p$ against the groundtruth target paper $y$. To ensure objective evaluation of scientific merit, we implement a structured comparison protocol:
\begin{align}
    r, J = \text{PaperReview}(\text{RandomSwap}(p, y); g)
\end{align}
This formulation employs a calibrated paper review agent that produces a comparative rating $r \in \{-3, -2, -1, 0, 1, 2, 3\}$, where positive values indicate the AI-generated paper exceeds the target paper in scientific contribution, zero indicates equivalence, and negative values signal inferior quality. The magnitude of $r$ quantifies the degree of scientific advancement or regression. The review agent also provides $J$, a structured set of justifications based on reviewing guidelines $g$ derived from ICLR conference standards—widely recognized in the machine learning community for emphasizing originality, technical soundness, and significance of contributions.

To ensure methodological rigor, we incorporate two critical debiasing mechanisms: (1) randomly swapping the presentation order of papers to eliminate position bias, and (2) conducting multiple independent evaluations using diverse state-of-the-art LLMs (including multiple GPT, Claude, and Gemini models) with temperature set to 1, creating a comprehensive panel-like review process that effectively mitigates individual model biases and enhances evaluation reliability. This carefully designed approach establishes a robust framework for quantifying whether AI systems can independently discover scientific insights that match or exceed those produced by human researchers.

\noindent \textbf{Alignment with Top-Tier Peer-Review Standards}. To ensure our evaluation framework upholds rigorous academic standards, we align Scientific Contribution Evaluation with established peer-review protocols from premier venues. Specifically, we assess research quality across critical dimensions, including technical novelty, methodological rigor, empirical validation, and impact--directly mirroring comprehensive evaluation criteria used in the ICLR conference review process.

To validate the reliability of our LLM-based evaluation mechanism, we conducted extensive benchmark experiments on a diverse sample of previously published ICLR papers with known acceptance decisions, demonstrating strong correlation between our automated assessments and the judgments rendered by expert human reviewers in real-world academic settings. Experiments using 5 popular LLMs with 64 randomly sampled ICLR submissions—forming 32 paper pairs for comparison—demonstrate that our LLM-based reviewer judgments perfectly align with ICLR's final decisions, in pairwise paper reviewing and confirming robust alignment with top-tier peer-review standards.

\section{The \model\ Framework}

\label{sec:solution}
\begin{figure*}[h]
    \centering
    \includegraphics[width=1\linewidth]{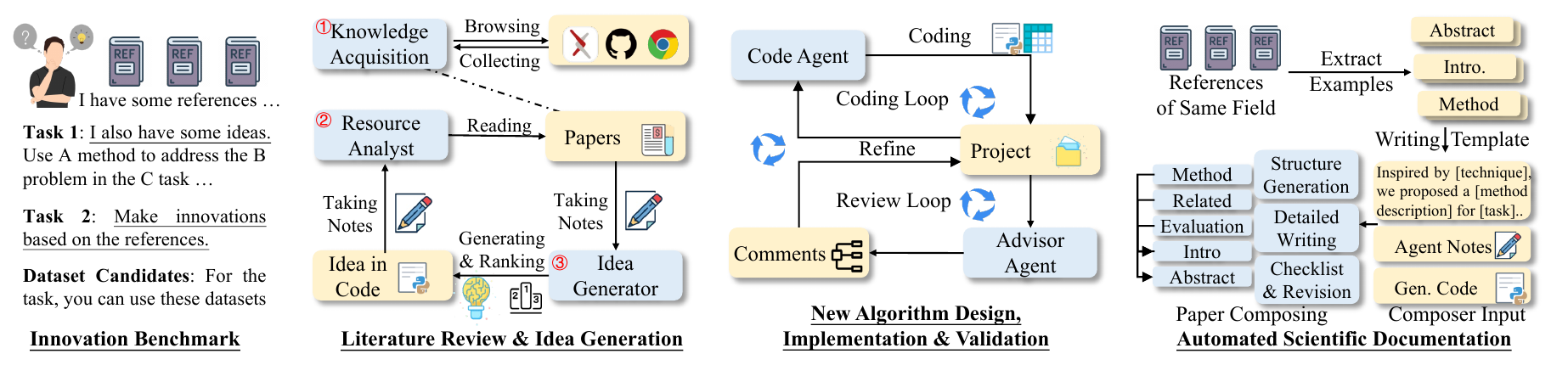}
    \caption{Architectural framework of \model: A comprehensive system of fully-automated LLM agents for end-to-end scientific discovery—seamlessly orchestrating literature review, idea generation, algorithm implementation, experimental validation, and paper writing.}
    \label{fig:overall}
\end{figure*}

\subsection{Multi-Agent System Overview of \model}
{Recent work has demonstrated the potential of end-to-end AI systems that autonomously generate scientific ideas, conduct experiments, and produce written documentation~\cite{lu2024aiscientist,yamada2025aiscientistv2}. Building on this paradigm, }
\model\ presents a comprehensive framework for autonomous research that systematically progresses through three key stages: \textbf{i) Literature Review and Idea Generation}; \textbf{ii) New Algorithm Design, Implementation and Validation} and \textbf{iii) Automated Scientific Documentation}. As illustrated in Figure~\ref{fig:overall}, this end-to-end research pipeline transforms initial scientific concepts into fully-developed academic contributions with rigorous methodology and minimal human intervention.

\subsubsection{Literature Review}

\noindent $\bullet$ \textbf{Knowledge Acquisition Agent}. The autonomous research process begins with the intensive literature exploration, driven by the \ag{Knowledge Acquisition Agent}. This specialized agent systematically discovers and extracts \emph{Relevant Papers} and \emph{Code Repositories} from scientific databases, establishing a solid foundation for the entire research pipeline.

A key advantage of AI-Researcher is its minimal input requirement—users need only provide 10-15 reference papers. The system then intelligently filters and processes this input to identify the most valuable information, significantly reducing the need for extensive datasets or manually gathered research files. Within this agent, \model\ performs two critical functions:

\noindent 1) \textbf{Code Repository Selection}: Using the user-provided reference papers as guidance, the agent applies sophisticated filtering algorithms to identify at least 5 high-quality GitHub repositories. This comprehensive filtering process meticulously evaluates multiple quality dimensions across diverse implementation paradigms: $\bullet$ \textbf{Code Recency} (prioritizing up-to-date implementations); $\bullet$ \textbf{GitHub Popularity} (star count as quality indicator); $\bullet$ \textbf{Documentation Quality} (completeness of README files); $\bullet$ \textbf{Domain Relevance} (alignment with research focus); $\bullet$ \textbf{Citation Impact} (scholarly influence).

\noindent 2) \textbf{Supplementary Literature Gathering}: For each filtered high-quality repository, the agent automatically retrieves corresponding papers from arXiv, including their complete LaTeX source files, further enriching the knowledge base with contextually relevant technical materials.

All operations execute within a secure containerized Docker environment, ensuring system integrity while enabling scalable computation across diverse research domains. The intelligent filtering and selection criteria guarantee that only the most relevant, maintained, and impactful resources form the knowledge foundation for subsequent research stages. Detailed system prompts and implementation tools for the \ag{Knowledge Acquisition Agent} are provided in the Appendix.

\noindent $\bullet$ \textbf{Resource Analyst Agent}. This specialized agent systematically deconstructs complex research concepts into manageable atomic components, meticulously extracting their mathematical formulations and corresponding code implementations through its dedicated \textbf{Paper Analyst} and \textbf{Code Analyst} sub-agents, ensuring precise alignment between theoretical expressions and practical implementation.

\noindent\textbf{Secure Research Environment.}
To protect host systems during automated operations, all analysis processes execute within a Docker containerized environment. This containment strategy offers three key advantages: (1) it establishes robust security boundaries that prevent unauthorized system modifications; (2) it provides consistent pre-configured environments with essential machine learning frameworks like PyTorch; and (3) it enables dynamic package management where agents can autonomously install additional dependencies as research needs evolve. The containerization approach creates a controlled yet flexible research workspace that safely supports the entire pipeline from literature analysis through implementation to manuscript generation.

\noindent \textbf{Integrated Research Analysis}. \ag{Resource Analyst} forms a critical bridge between abstract concepts and their concrete implementations, significantly reducing potential hallucinations in subsequent development stages. This agent operates through a carefully structured analytical process:

\noindent 1) \textbf{Concept Decomposition}: Using the initial research idea as conceptual guidance and foundation, the agent methodically decomposes complex research objectives into atomic academic concepts--fundamental, indivisible research elements requiring individual investigation.

\noindent 2) \textbf{Mathematical Formalization}: The \ag{Paper Analyst} examines downloaded LaTeX files through RAG-based paradigm, systematically extracting mathematical formulations of each atomic concept. This creates a formal foundation for all subsequent implementation steps.

\noindent 3) \textbf{Implementation Analysis}: The \ag{Code Analyst} sub-agent then meticulously analyzes the downloaded code repositories to locate corresponding implementations of these mathematical expressions, identifying critical reference files and associated dependencies.

\noindent 4) \textbf{Knowledge Integration}: Results from both paper and implementation analyses are meticulously synthesized into comprehensive concept profiles, effectively establishing clear bidirectional connections between mathematical formulations and their practical code implementations.

This rigorous cycle continues until all concepts are thoroughly investigated, culminating in a detailed research report that serves as the foundation for development planning. The \ag{Plan Agent} transforms these findings into a comprehensive implementation roadmap addressing training procedures, testing methodologies, and dataset requirements--creating a complete, executable research strategy.

\subsubsection{Idea Generation}

{Scientific discovery often emerges where deep domain expertise intersects with bold, creative inquiry. Recent advances in LLMs have shown promise in in assisting research ideation. Systems like Chain-of-Ideas~\cite{li2024chain} organize literature in progressive chains to mirror knowledge evolution, while ResearchAgent~\cite{baek2025researchagent} introduces collaborative LLM reviewers to iteratively refine research proposals. Despite their demonstrated effectiveness, these systems often remain anchored in the recombination or reinterpretation of known knowledge. In contrast, our \ag{Idea Generator} is expressly designed to venture beyond established paradigms, systematically targeting the conceptual frontiers of science.}

Operating after comprehensive theoretical and empirical analysis, the \ag{Idea Generator} employs sophisticated knowledge synthesis techniques to identify unexplored research territories. The agent deliberately seeks conceptual gaps, contradictory findings, and emerging patterns across literature and implementations—precisely the spaces where scientific discoveries often emerge. Each generated proposal deliberately pushes beyond established paradigms through:

\begin{itemize}[leftmargin=*]
    \item \textbf{\emph{Challenges}}: pinpoint fundamental limitations in current scientific understanding
    \item \textbf{\emph{Existing Methods}}: analysis revealing conceptual blind spots ripe for innovation
    \item \textbf{\emph{Motivation}}: establishing scientific necessity for paradigm-shifting approaches
    \item \textbf{\emph{Proposed Method}}: introducing novel theoretical frameworks or algorithmic innovations
    \item \textbf{\emph{Technical Details}} translating abstract breakthroughs into implementable science
    \item \textbf{\emph{Expected Outcomes}}: projecting potential scientific and practical impact
\end{itemize}

\textbf{Divergent-Convergent Discovery Framework:} Our discovery process employs a multi-phase approach for scientific originality. Inspired by~\cite{idea_agent}, the divergent phase generates five conceptually distinct research directions, exploring orthogonal perspectives and cross-disciplinary connections. These undergo rigorous convergent evaluation against criteria including \emph{Scientific Novelty}, \emph{Technical Soundness}, and \emph{Transformative Potential}. The most promising concept receives comprehensive development, resulting in a proposal that charts new scientific territory with clear implementation pathways.


\subsection{New Algorithm Design, Implementation and Validation}
Translating novel research concepts into functioning implementations represents one of the most challenging aspects of computational science. Unlike traditional code agents that attempt one-shot implementations--often causing errors or research misalignment--we introduce a framework that mirrors the proven human research paradigm of iterative refinement and collaborative feedback.

\subsubsection{Multi-Stage Refinement Architecture} 
Our approach implements a cyclical development process with explicit feedback mechanisms, enabling progressive algorithm improvement through multiple refinement cycles. This methodology not only increases implementation success rates but also allows for test-time scaling—similar to how researchers refine their work through extended collaborative iterations. The framework deliberately models the advisor-student relationship that underlies successful academic research, providing structured guidance while maintaining implementation flexibility.

\noindent $\bullet$ \textbf{Multi-Stage Refinement Architecture.} 
Our approach implements a cyclical development process with explicit feedback mechanisms, enabling progressive algorithm improvement through multiple refinement cycles. This methodology not only increases implementation success rates but also allows for test-time scaling—similar to how researchers refine their work through extended collaborative iterations. The framework deliberately models the advisor-student relationship that underlies successful academic research, providing structured guidance while maintaining implementation flexibility.

\noindent $\bullet$ \textbf{Code Implementation Framework.} 
The \ag{Code Agent} transforms research analysis and development plans into executable implementations across diverse domains. Operating within a controlled workspace, this agent creates structured implementations with comprehensive file system and execution capabilities. It enforces strict code independence principles while ensuring faithful translation of academic concepts into working code. Throughout development, the agent maintains continuous verification against the implementation plan with thorough documentation of all modifications.

\noindent $\bullet$ \textbf{Expert Validation Framework.}
Our \ag{Advisor Agent} provides expert feedback that bridges the gap between theoretical concepts and practical implementation. It validates implementation fidelity by systematically comparing code against atomic research ideas extracted during analysis. The agent examines results through specialized navigation tools and visualizations while referencing workspace materials. Based on comprehensive evaluation, it generates detailed assessment reports with specific, actionable modification recommendations to guide refinement iterations.

\noindent $\bullet$ \textbf{Progressive Experimental Cycles.}
Our experimental process implements a rigorous scientific approach to code validation. The \ag{Code Agent} begins by developing prototype implementations that undergo initial testing on minimal data (typically 1-2 epochs or small dataset subsets) to establish baseline feasibility. Following this preliminary validation, successful implementations that incorporate review feedback advance to full-scale experiments, while persistently unsuccessful implementations receive "unfeasible" classification after multiple refinement attempts. Throughout this cyclical process, the \ag{Advisor Agent} provides analytical support by evaluating results and recommending supplementary investigations. These recommendations span implementation refinements, validation studies, visualizations, and comparative analyses aligned with established research standards. Through these structured refinement cycles, implementations systematically evolve toward optimal performance, ensuring scientific rigor and reproducibility in our findings.

\begin{wrapfigure}{r}{0.63\linewidth}
    \centering
    \vspace{-0.25in}
    \includegraphics[width=\linewidth]{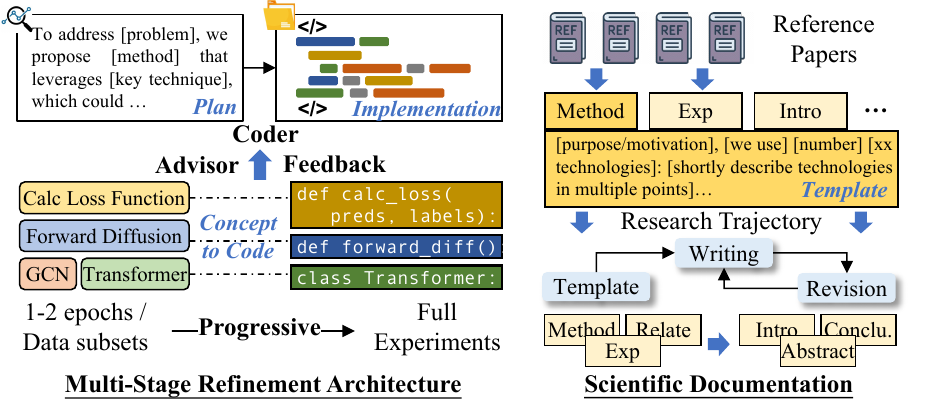}
    \vspace{-0.22in}
    \caption{Illustration of (1) multi-stage implementation refinement, and (2) automated sceintfic documentation.}
    \label{fig:technical_modules}
    \vspace{-0.18in}
\end{wrapfigure}

\subsection{Automated Scientific Documentation}

The culmination of scientific research requires transforming raw experimental results into structured academic knowledge contributions. Following substantial implementation and experimentation cycles, our \ag{Documentation Agent} initiates a sophisticated process that converts technical artifacts into publication-ready manuscripts while maintaining scientific integrity and narrative coherence.

\noindent\textbf{Research Trajectory Synthesis.}
The \ag{Automated Documentation Agent} systematically integrates diverse research elements—including agent reasoning processes, execution logs, implemented code, and experimental outcomes—into cohesive scientific narratives. This holistic approach preserves the complete intellectual context behind discoveries while structuring findings according to established academic conventions. Unlike simple documentation tools, our system captures both the final results and the critical decision pathways that led to scientific advances.

\noindent\textbf{Overcoming Document-Scale Coherence Challenges.}
Academic manuscripts require sustained coherence across thousands of words--a significant challenge for LLMs that typically struggle with cross-reference consistency and factual stability over extended outputs. Analyzing how researchers draft papers progressively from outlines to full texts and inspired by~\cite{storm}, we developed a multi-stage generation framework that mirrors this hierarchical approach. This methodology systematically overcomes LLM limitations by decomposing the complex writing task into manageable components while preserving logical connections across sections and maintaining factual integrity throughout the document.

\noindent\textbf{Three-Phase Hierarchical Documentation.} 
Our writing approach employs a systematic three-stage process: (1) \textbf{Synthesizing Research Artifacts}: structural outlining based on domain-appropriate templates that establish section hierarchies and logical flow; (2) \textbf{Template-Guided Structure}: methodical content elaboration that develops explanations maintaining cross-document consistency; and (3) \textbf{Hierarchical Documentation Process}: systematic verification using specialized academic checklists that identify and remediate inaccuracies or omissions. This "one more step" review process enhances factual integrity and completeness, ensuring manuscripts meet publication standards without the hallucinations and inconsistencies that typically plague LLM-generated long-form content.

\section{Experiments}
\label{sec:eval}

Our evaluation addresses six key research questions: \textbf{RQ1}: How complete and correct are \model's methodology implementations? \textbf{RQ2}: How does AI-generated research compare to groundtruth human research? \textbf{RQ3}: What is AI-Researcher's capability in conducting open-ended scientific exploration? \textbf{RQ4}: How does using different LLMs impact the performance of our AI-Researcher? \textbf{RQ5}: How well does the automated paper review agent align with expert peer-review assessments? \textbf{RQ6}: In which specific aspects does AI-Researcher's research match or exceed human research quality? The following sections provide detailed answers to these questions.

\subsection{Experimental Settings}
\begin{table}[t]
    \centering
    \caption{Data statistics of Scientist-Bench across diverse research domains, featuring comprehensive task distribution across guided innovation and open-ended exploration challenges.}
    \setlength{\tabcolsep}{1.6mm}
    \label{tab:bench}
    \begin{tabular}{lcccc}
        \toprule
        \textbf{Research Domain} & \textbf{\# Papers} & \textbf{\# Level-1} & \textbf{\# Level-2} & \textbf{\# Rejected Papers}\\
        \midrule
        Diffusion Models & 4 & 4 & 1 & 0\\
        Vector Quantization & 6 & 6 & 1 & 0\\
        Graph Neural Networks & 7 & 7 & 1 & 1\\
        Recommender Systems & 5 & 5 & 3 & 1\\
        \midrule
        Total & 22 & 22 & 6 & 2\\
        \bottomrule
    \end{tabular}
\end{table}

\noindent\textbf{Experimental Dataset: Benchmarking Scientific Innovation}. We evaluate our \model\ using the \bench\ benchmark (as described in Section 2)--a meticulously curated collection of 22 state-of-the-art papers spanning several critical AI domains including Computer Vision (\eg, Diffusion Model), Signal Processing (\eg, Vector Quantization), Graph Learning (\eg, Graph Neural Networks), and Information Retrieval (\eg, Recommender Systems). Our work addresses a significant gap in the field, as comprehensive benchmarks for scientific innovation assessment remain notably scarce. The evaluation protocol employs \textbf{Two Complementary Innovation Tasks of Varying Difficulty Levels} (detailed below) designed to test distinct research capabilities across diverse methodological paradigms. Table~\ref{tab:bench} presents the complete dataset statistics, establishing important baseline measures in this underexplored evaluation landscape for future comparative analyses.

\begin{itemize}[leftmargin=*]

    \item \textbf{Level-1 Task: Guided Innovation} — The scientific discovery agent receives explicit research instructions alongside reference papers, simulating scenarios where researchers pursue specific innovation targets. This structured evaluation provides clear assessment of the agent's ability to develop targeted innovations and spans all 22 groundtruth papers for comprehensive coverage.\\\vspace{-0.12in}

    \item \textbf{Level-2 Task: Autonomous Exploration} — The scientific discovery agent performs independent, open-ended research exploration with only reference papers as input. This more challenging scenario tests the agent's capacity to identify promising research gaps and generate novel directions without explicit guidance—a crucial capability for truly autonomous scientific assistants. To maintain methodological integrity and prevent cross-contamination from overlapping reference materials, we strategically selected 5 groundtruth papers across distinct research domains, enabling us to rigorously assess genuine discovery abilities without confounding influences.

\end{itemize}


\noindent \textbf{Evaluation Protocols}. To assess scientific contributions of \model, we implement a two-stage evaluation framework examining both technical implementation and research quality:\vspace{-0.05in}

\begin{itemize}[leftmargin=*]

    \item \textbf{Implementation Verification}. We employ a specialized code review agent to verify that AI-generated code faithfully implements the proposed methodology described in the technical report. This critical validation step ensures the practical reproducibility of the research contribution. We quantify performance using the completion ratio $R$, defined as the fraction of AI implementations correctly executing the intended research approach. This metric directly measures the model's ability to translate conceptual innovations into functional implementations.\\\vspace{-0.12in}

    \item \textbf{Scientific Quality Assessment}. For implementations that successfully pass verification, we perform an in-depth comparative analysis between AI-generated research and their human-authored counterparts. This evaluation mirrors the rigorous peer-review process typical at prestigious venues like ICLR and NeurIPS~\cite{jin2024agentreview}, where an expert review agent systematically examines each paper pair through three fundamental scientific dimensions: $\bullet$ (1) innovation and novelty of research contributions, $\bullet$ (2) theoretical and methodological rigor, and $\bullet$ (3) empirical validation and experimental design quality. This approach ensures our assessment adheres to established standards of scientific excellence in the field. The evaluation culminates in a comprehensive comparative rating on a 7-point scale (-3 to +3), where negative scores indicate AI work falling below human standards, zero represents parity, and positive scores signify AI research exceeding human benchmarks. Each rating is substantiated with detailed justifications citing specific evidence from both papers, providing transparent rationale for the comparative assessment.

\end{itemize}

\noindent \textbf{LLMs as Judges with Robust Evaluation}. To establish robust evaluation, we leverage five state-of-the-art LLMs (GPT-4, o1-mini, o3-mini, Claude-sonnet-3.5, and Claude-sonnet-3.7), each performing 16 independent assessments per paper with temperature=1.0. This ensemble approach mitigates individual model biases and provides statistical confidence in our findings. We analyze results through two complementary metrics: (1) mean rating across all evaluations—quantifying the quality gap between AI and human research, and (2) percentage of AI papers scoring $\geq$-1.0—representing research contributions that achieve at least near-human quality in the field.

\subsection{Dual-Metric Evaluation Framework: Quantifying Implementation Quality (RQ1)}
\label{sec:eval_code}
To evaluate the stability and quality of \model\ system's code implementation based on requirements, we propose \textbf{Completeness} and \textbf{Correctness} metrics for measurement. 

\begin{wrapfigure}{l}{0.5\textwidth} 
  \centering
  \includegraphics[width=0.45\textwidth]{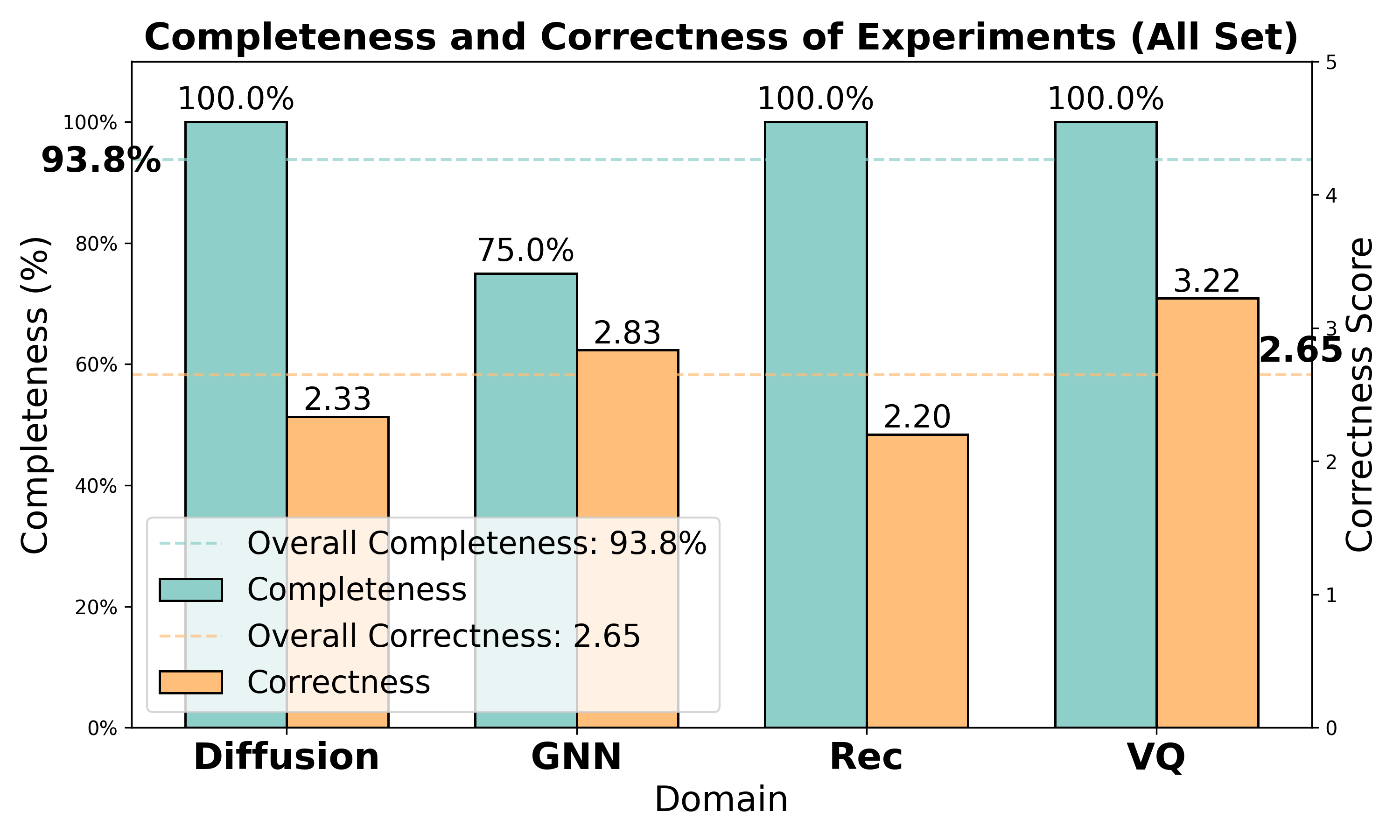}
  \caption{Quantifying Implementation Quality in terms of Completeness and Correctness.}
  \label{fig:all_set}
\end{wrapfigure}

Specifically, we evaluate implementation quality across two critical dimensions: $\bullet$ \textbf{Completeness} measures whether the agent produces executable code within the allocated inference budget. We implement an unambiguous termination protocol where agents signal success via \texttt{case\_resolved} or acknowledge failure through \texttt{case\_not\_resolved}, enabling automated assessment of task completion rates. $\bullet$ \textbf{Correctness} addresses a nuanced challenge--even when code executes, it may contain subtle implementation flaws, conceptual misalignments, or missing components requiring deeper analysis. To evaluate implementation fidelity, we employ a multi-agent framework where an \ag{Advisor Agent} generates detailed analysis reports identifying potential issues, followed by a \ag{Judge Agent} that assigns quality scores on a 5-point scale. The final correctness metric represents the mean score across multiple independent judgments, providing a robust measure of implementation quality throughout the development lifecycle. We conduct extensive evaluations across both Level 1 and Level 2 tasks in our benchmark, systematically analyzing how implementation performance varies with different backbone LLMs. Our analysis reveals several key findings:


\noindent\textbf{Performance Analysis}. 
We conducted comprehensive experiments using Claude-series models across our entire benchmark dataset, evaluating both completeness and correctness metrics as shown in Figure~\ref{fig:all_set}. The results reveal remarkable stability--our \model\ system achieves an outstanding 93.8\% completeness rate with Claude-series models, failing only in cases involving complex technical challenges such as tensor dimension conflicts and datatype mismatches that persisted despite multiple debugging iterations. This exceptional completeness rate underscores the robustness of our system's implementation and debugging capabilities across diverse computational and algorithmic domains.

For correctness, our system achieves an average score of 2.65 (on a 1-5 scale), exceeding the median threshold and indicating successful implementation of the majority of specified requirements. Notably, performance varies across domains—Vision and Question Answering (VQ) tasks reached the highest correctness of 3.22, while Recommendation (Rec) tasks averaged 2.20. This variation likely reflects the inherent complexity differences between domains, with recommendation systems typically requiring more intricate algorithmic implementations and data handling procedures.

\begin{figure*}[htbp]
    \centering
    \begin{minipage}[b]{0.49\textwidth}
        \centering
        \includegraphics[width=\textwidth]{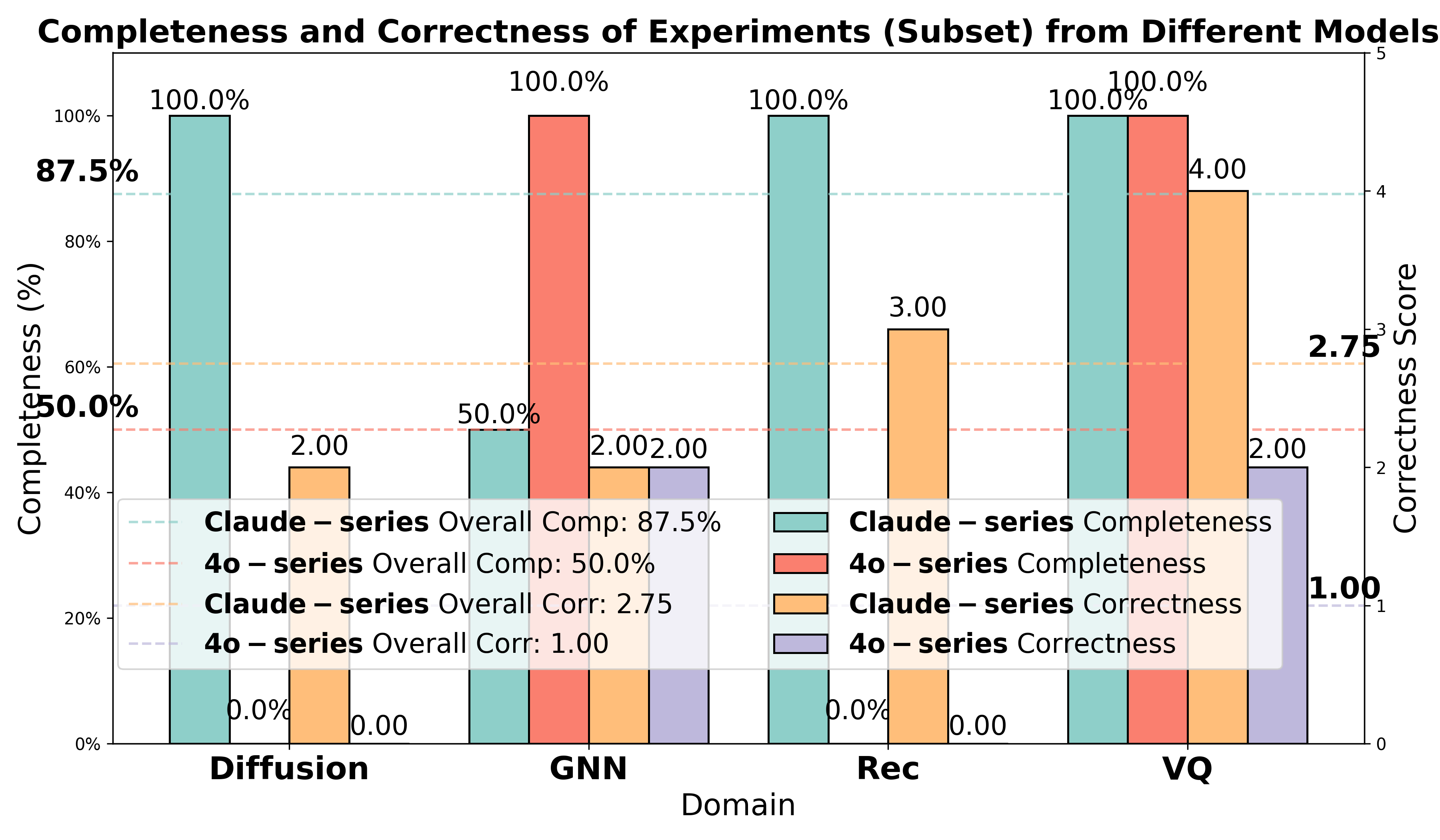}
        \label{fig:4o}
    \end{minipage}
    \hfill
    \begin{minipage}[b]{0.49\textwidth}
        \centering
        \includegraphics[width=\textwidth]{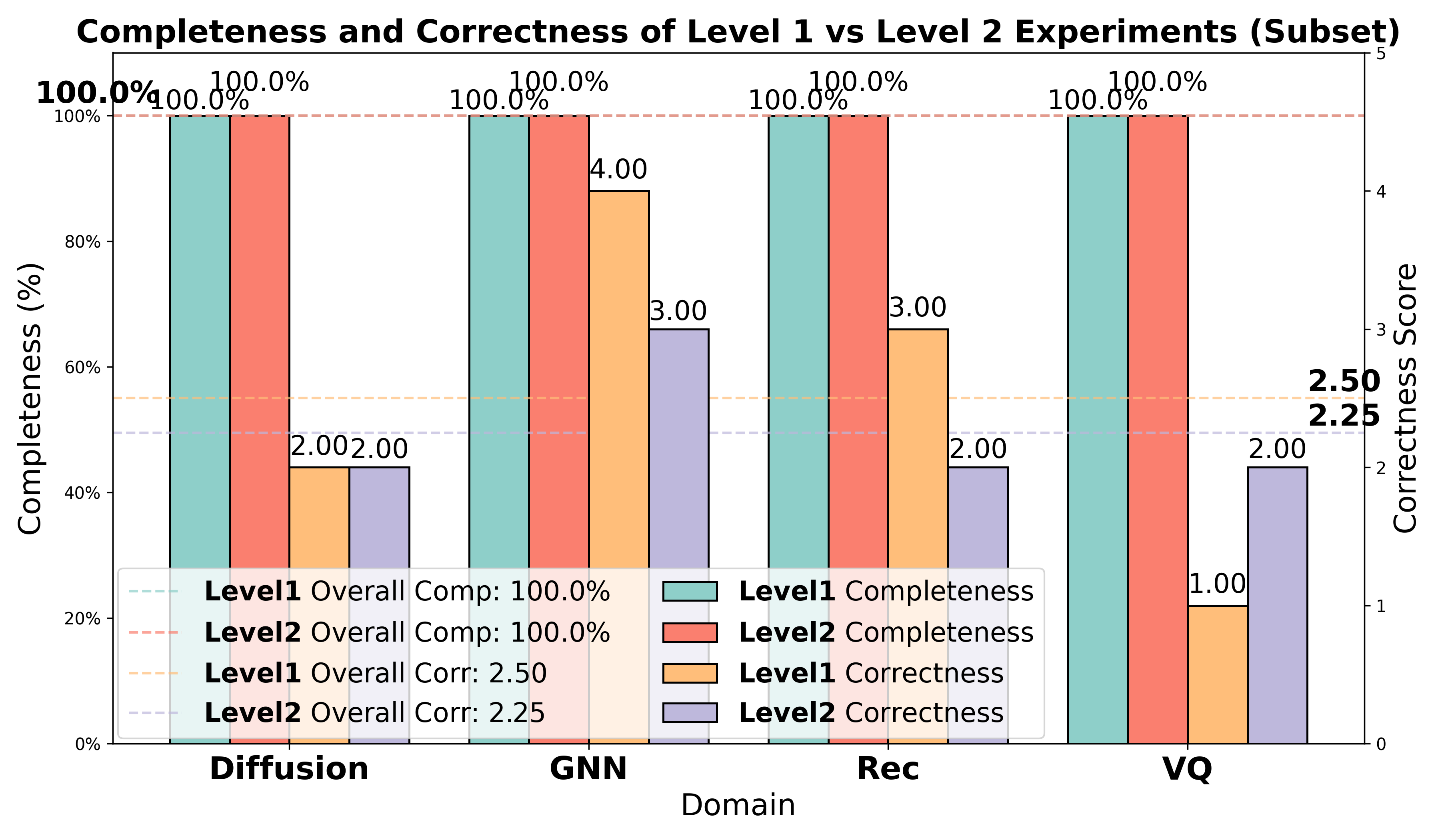}
        \label{fig:idea}
    \end{minipage}
    \caption{Performance Comparison Across Model Families and Task Complexity. Left: Claude-series versus 4o-series models on implementation completeness and correctness metrics (benchmark subset). Right: Claude-series performance across Level 1 (adaptation) and Level 2 (innovation) tasks.}
    \label{fig:performance_comparison}
\end{figure*}


\noindent\textbf{Performance Comparison between LLMs in Scientific Implementation}.
To rigorously compare the capabilities of different large language models in automated scientific research, we conducted a controlled evaluation using a balanced subset of our benchmark dataset spanning multiple technical domains. As illustrated in Figure~\ref{fig:performance_comparison} (left), our assessment reveals substantial performance differences between model families. Claude-series models achieved an impressive 87.5\% completeness rate on the evaluation subset, significantly outperforming the 4o-series models which reached only 50\% completeness. This performance gap stems primarily from differences in debugging proficiency—the 4o-series models frequently generated code with persistent tensor dimension mismatches and training instabilities (NaN losses) that remained unresolved despite multiple debugging attempts. In contrast, Claude-series models demonstrated superior problem-solving capabilities, successfully identifying and resolving complex implementation issues through systematic debugging approaches.

The quality disparity extends beyond mere code completion to implementation correctness, where Claude-series models scored substantially higher (2.75 points average) compared to 4o-series models (1.0 point average). The 4o-series implementations exhibited a consistent pattern of oversimplification and conceptual omissions in complex tasks. A particularly illustrative example occurred in the diffusion model integration task, where the 4o-series model claimed successful implementation of a Diffusion Transformer architecture while detailed inspection revealed merely a standard Vision Transformer (ViT) implementation with complete absence of the critical diffusion components. This systematic evaluation highlights the importance of both implementation completeness and conceptual correctness when assessing LLM capabilities for advanced scientific research tasks.


\noindent\textbf{Implementation Success with Increasing Task Complexity (Level-2)}.
To systematically evaluate our framework's performance across difficulty levels, we conducted a comparative analysis using a balanced subset of benchmark tasks from each research domain. Figure~\ref{fig:performance_comparison} (right) presents the completeness and correctness metrics for Level 1 tasks (adapting established methodologies) versus Level 2 tasks (generating and implementing novel research ideas) using Claude-series models.

Remarkably, \model\ maintains perfect implementation completeness (100\%) even for the more challenging Level 2 innovation tasks. This consistency demonstrates the robustness of our system's self-debugging mechanisms and execution pipeline when handling both established and novel methodological approaches. However, we observe a modest decrease in correctness scores from Level 1 (2.5) to Level 2 (2.25) tasks. This slight performance gap reveals an important challenge: while \model\ can reliably execute self-generated research ideas to completion, the implementation quality of novel concepts occasionally falls short of adaptation tasks.

The correctness differential stems primarily from two factors. First, the complexity of agent-generated research ideas varies considerably, with some innovations proving technically challenging to implement correctly. Second, while our idea generation and ranking system generally produces feasible concepts, the framework occasionally struggles to perfectly realize ambitious or complex innovations. These findings suggest promising avenues for future enhancement, particularly in developing sophisticated idea feasibility assessment mechanisms and implementing adaptive modification capabilities that allow real-time refinement of research approaches during implementation when obstacles arise.

\subsection{Evaluating Scientific Quality Through Pairwise Comparison ((RQ2)}
To rigorously assess the scientific merit of research generated by our \model\ framework, we implemented a systematic pairwise evaluation protocol comparing AI-generated papers against corresponding human-authored publications in the same research domains. Our evaluation methodology employs specialized paper review agents that perform detailed comparative analyses following established academic conference standards. These agents generate comprehensive reviews for each paper pair using ICLR reviewing guidelines—evaluating the clarity of research motivation, methodological soundness, technical innovation, and experimental validation across both works.
\begin{table}[t]
    \centering
    \caption{Comparative Evaluation of AI-Generated versus Human Research. Results show performance across four domains (Diffusion Models, Vector Quantization, Graph Neural Networks, and Recommender Systems) using two metrics: Mean Rating (-3=significantly inferior to 3=significantly superior) and Comparable Percentage (proportion of AI-generated papers receiving ratings $\geq$ -1.0). Evaluation using multiple independent LLM judges demonstrates domain-specific variations in AI research quality relative to established human benchmarks through comprehensive assessment criteria.}
    \setlength{\tabcolsep}{1.6mm}
    \label{tab:pairwise}
    \begin{tabular}{lcccccc}
        \toprule
        \textbf{Field} & \textbf{Metric} & \textbf{GPT-4o} & \textbf{o1-mini} & \textbf{o3-mini} & \textbf{Claude-3.5} & \textbf{Claude-3.7}\\
        \midrule
        \multirow{2}{*}{Diffusion} & Average Rating & -0.48$\pm$0.87 & -1.36$\pm$1.41 & -1.27$\pm$0.91 & -1.83$\pm$0.88 & -1.49$\pm$1.49\\
        & Comparable(\%) & 75.00\% & 25.00\% & 50.00\% & 0.00\% & 25.00\%\\
        \midrule
        \multirow{2}{*}{VQ} & Average Rating & -0.55$\pm$1.00 & -0.95$\pm$1.56 & -1.49$\pm$0.66 & -1.68$\pm$1.37 & -2.11$\pm$1.21\\
        & Comparable(\%) & 83.33\% & 50.00\% & 16.67\% & 16.67\% & 0.00\%\\
        \midrule
        \multirow{2}{*}{GNN} & Average Rating & -0.70$\pm$1.10 & -1.52$\pm$1.30 & -1.68$\pm$0.62 & -1.86$\pm$0.86 & -1.83$\pm$1.41\\
        & Comparable(\%) & 71.43\% & 42.86\% & 0.00\% & 0.00\% & 14.29\%\\
        \midrule
        \multirow{2}{*}{Rec} & Average Rating & -0.33$\pm$0.91 & -0.42$\pm$0.86 & -1.50$\pm$0.94 & -0.88$\pm$1.62 & -0.81$\pm$1.76\\
        & Comparable(\%) & 100.00\% & 100.00\% & 0.00\% & 40.00\% & 60.00\%\\
        \midrule
        \multirow{2}{*}{Overall} & Average Rating & -0.53$\pm$1.00 & -1.09$\pm$1.60 & -1.51$\pm$0.78 & -1.58$\pm$1.28 & -1.70$\pm$1.54\\
        & Comparable(\%) & 81.82\% & 54.55\% & 13.64\% & 13.64\% & 22.73\%\\
        \bottomrule
    \end{tabular}
\end{table}

\noindent $\bullet$ \textbf{Overall Performance}. Comparative evaluation reveals that while papers generated by \model\ receive moderately lower average ratings than human-authored works (ranging from -0.58 to -1.76 across evaluators), a substantial proportion of AI-generated papers (15.79\% to 78.95\%) demonstrate quality comparable to human research. This finding is particularly significant considering our benchmark comprises exclusively top-tier human-authored publications carefully selected from leading venues in each domain. The results demonstrate \model's remarkable capacity to execute the complete scientific research pipeline—from developing methodologically sound technical innovations to conducting rigorous experimental validations and synthesizing findings into coherent, well-structured academic manuscripts that approach quality standards of expert human researchers.

\noindent $\bullet$ \textbf{LLM Evaluator Divergence}. GPT-4o provides the highest ratings for AI-generated papers (78.95\% comparable with average rating -0.58), while Claude-3.7 gives the lowest ratings on average (21.05\% comparable with average rating -1.76). Moreover, for different research fields, LLM evaluators show varying preferences. For example, GPT-4o and o1-mini consider all generated recommendation papers comparable to groundtruth human papers, while o3-mini rates them as inferior. This demonstrates the potential bias of using only one LLM evaluator to assess the generated research works. In summary, different LLM evaluators yield varying comparable percentages, ranging from 15.79\% to 78.95\%, demonstrating that the AI-conducted research approaches the quality of top-tier human research.

\noindent $\bullet$ \textbf{Domain-Specific Analysis}. Performance varies across research fields but shows no consistent patterns. Papers on diffusion models gain higher comparable rate compared to GNN papers when evaluated with GPT-4o and Claude-3.7. However, this situation is reversed when using o1-mini as the evaluator. Recommendation papers achieves high comparable rate across all evaluators except o3-mini, while o3-mini thinks none of the generated recommendation papers are comparable to human papers. For the vector quantization domain, three evaluators (GPT-4o, o1-mini, Claude-3.5) think the generated papers are better than diffusion papers, while o3-mini and Claude-3.7 consider them worse but diffusion papers better. These variations appear to be more influenced by evaluator preferences than by domains, suggesting that \model\ maintains consistent performance across different research domains without catastrophic degradation in any particular field.

\subsection{Open-Ended Autonomous Scientific Innovation Capabilities (RQ3)}
To assess \model's capacity for genuine scientific innovation, we evaluated its performance on open-ended research tasks (level-2) where the system receives only reference literature without explicit research directives. This challenging scenario requires \model\ to independently identify promising research directions, formulate novel hypotheses, and execute the complete scientific workflow from conceptualization through implementation and documentation. We systematically evaluated the resulting technical manuscripts against established human-authored publications using our specialized paper review framework. Table~\ref{tab:open-ended} presents a comprehensive analysis of these autonomous research outputs across diverse domains.

For this evaluation, we carefully selected 5 representative papers spanning distinct research areas to ensure methodological diversity while accounting for the natural citation overlap within specialized research communities. Our analysis of these autonomous scientific explorations reveals several key insights into the system's creative research capabilities:

\begin{table}[t]
    \centering
    \caption{Results of open-ended research exploration evaluated by different LLMs. Mean Rating ranges from -3 (significantly inferior) to 3 (significantly superior). Comparable Papers shows the percentage of AI-generated papers rated above -1.0 compared to groundtruth papers.}
    \setlength{\tabcolsep}{1.6mm}
    \label{tab:open-ended}
    \begin{tabular}{lcccccc}
        \toprule
        \textbf{Field} & \textbf{Metric} & \textbf{GPT-4o} & \textbf{o1-mini} & \textbf{o3-mini} & \textbf{Claude-3.5} & \textbf{Claude-3.7}\\
        \midrule
        \multirow{2}{*}{Diffusion} & Average Rating & -0.56$\pm$0.79 & -1.75$\pm$0.83 & -1.00$\pm$0.50 & -2.00$\pm$0.00 & -0.56$\pm$1.41 \\
        & Comparable(\%) & 100.00\% & 0.00\% & 100.00\% & 0.00\% & 100.00\%\\
        \midrule
        \multirow{2}{*}{VQ} & Average Rating & -0.25$\pm$0.97 & -0.62$\pm$0.99 & -0.88$\pm$0.99 & -1.00$\pm$1.50 & -1.31$\pm$1.10\\
        & Comparable(\%) & 100.00\% & 100.00\% & 100.00\% & 100.00\% & 0.00\%\\
        \midrule
        \multirow{2}{*}{GNN} & Average Rating & 0.12$\pm$0.78 & -0.50$\pm$1.00 & -2.19$\pm$1.24 & -1.44$\pm$0.50 & -0.94$\pm$1.43\\
        & Comparable(\%) & 100.00\% & 100.00\% & 0.00\% & 0.00\% & 100.00\%\\
        \midrule
        \multirow{2}{*}{Rec} & Average Rating & 0.06$\pm$0.92 & -0.77$\pm$1.52 & -1.08$\pm$1.00 & 0.19$\pm$1.78 & -0.96$\pm$1.70\\
        & Comparable(\%) & 100.00\% & 66.67\% & 66.67\% & 100.00\% & 33.33\%\\
        \midrule
        \multirow{2}{*}{Overall} & Average Rating & -0.23$\pm$0.99 & -0.85$\pm$1.32 & -1.22$\pm$1.07 & -0.65$\pm$1.66 & -0.95$\pm$1.54\\
        & Comparable(\%) & 100.00\% & 66.67\% & 66.67\% & 66.67\% & 50.00\%\\
        \bottomrule
    \end{tabular}
\end{table}

\noindent $\bullet$ \textbf{Performance Analysis}. A striking pattern emerges when comparing \model's performance across task structures: the system demonstrates markedly superior outcomes in open-ended level-2 scenarios versus instruction-guided level-1 tasks. This quality differential manifests consistently across evaluation metrics, with average ratings improving substantially from -0.58~-1.76 to -0.20~-1.01, and comparable rates rising dramatically from 15.79\%~78.95\% to 40.00\%~100.00\%.

These findings challenge conventional assumptions about AI research capabilities, suggesting that \model\ excels when leveraging its internal knowledge synthesis and ideation processes rather than following explicit research directives. The notable performance enhancement indicates that prescriptive research instructions may inadvertently constrain the system's creative exploration capacity, while autonomous research formulation allows \model\ to identify and pursue more scientifically promising directions that better align with its implementation capabilities.



\noindent $\bullet$ \textbf{Domain-Specific Resource Constraints Influence Innovation Quality}. Our cross-domain analysis reveals a systematic relationship between computational requirements and autonomous research performance. Research areas with lighter computational demands, particularly recommender systems, demonstrate remarkable quality improvements in open-ended explorations, achieving impressive comparable rates of 66.67\%-100\% across most evaluator benchmarks.

Conversely, computationally intensive domains such as diffusion models exhibit more modest gains in evaluation metrics despite similar conceptual innovation. This consistent pattern suggests that \model's fundamental research capabilities extend beyond what our implementation currently demonstrates, with performance disparities reflecting practical resource limitations rather than conceptual understanding deficiencies. The finding highlights the importance of computational capacity as a determining factor in AI research quality, indicating substantial potential for enhanced performance should greater computational resources become available.

\subsection{Impact of LLM Backbones (RQ4)} 
To systematically isolate the influence of foundation model selection on research capabilities, we conducted controlled ablation studies across different LLM backbones while maintaining identical system architecture, research tasks, and evaluation protocols. We selected a representative set of 7 research problems spanning diverse domains to comprehensively assess model-specific performance variations. Table~\ref{tab:backbone} presents the comparative analysis results, revealing significant and consistent performance differentials between foundation models.

The empirical evidence demonstrates Claude-3.5's substantial advantage as the research agent backbone, with this configuration consistently achieving higher mean quality ratings across all evaluator benchmarks compared to GPT-4o implementations. This performance differential extends beyond simple metrics to comparable rates, where Claude-3.5 outperforms in most evaluation contexts, with the exception of o1-mini assessments. The quality gap becomes particularly pronounced under the most stringent evaluation criteria (o3-mini), where Claude-3.5-based systems produce research comparable to human standards while GPT-4o-based configurations fail to generate any research meeting minimum comparability thresholds. These findings highlight the critical importance of foundation model selection in determining the upper bounds of automated scientific research quality.

\begin{table}[t]
    \centering
    \caption{Comparative quality assessment between research papers generated using different LLM backbones versus human-authored benchmarks, with comprehensive evaluation across multiple independent reviewers revealing consistent performance differentials between foundation architectures.}
    \setlength{\tabcolsep}{1.4mm}
    \label{tab:backbone}
    \begin{tabular}{ccccccc}
        \toprule
        {\textbf{Research}} & {\textbf{Evaluation}} & \multicolumn{5}{c}{\textbf{LLM used in Reviewing Agent}}\\
        \cline{3-7}
        \textbf{Agent LLM} & \textbf{Metric} & \textbf{GPT-4o} & \textbf{o1-mini} & \textbf{o3-mini} & \textbf{Claude-3.5} & \textbf{Claude-3.7}\\
        \midrule
        \multirow{2}{*}{GPT-4o} & Average Rating & 0.69$\pm$1.05 & -1.45$\pm$1.40 & -1.62$\pm$0.55 & -2.05$\pm$0.23 & -2.12$\pm$1.11\\
        & Comparable(\%) & 71.43\% & 42.86\% & 0.00\% & 0.00\% & 14.29\%\\
        \midrule
        \multirow{2}{*}{Claude-3.5} & Average Rating & 0.59$\pm$1.01 & -1.42$\pm$1.43 & -1.44$\pm$0.72 & -1.80$\pm$1.03 & -1.98$\pm$1.45\\
        & Comparable(\%) & 85.71\% & 28.57\% & 14.29\% & 0.00\% & 28.57\%\\
        \bottomrule
    \end{tabular}
\end{table}

\subsection{Paper Review Agent Validation Against Human Expert Judgments (RQ5)}
To rigorously validate our automated review system's alignment with expert scientific assessment, we conducted a systematic evaluation using gold-standard human judgment data from the ICLR conference. We constructed a validation dataset comprising 32 carefully sampled paper pairs from proceedings (2021-2023), where each pair contains one accepted and one rejected submission. To ensure meaningful comparative analysis and maintain consistency with our main experimental protocol, we prioritized paper pairs exhibiting high TF-IDF similarity in content and focus.

We applied identical pairwise review methodology as our main experiments, evaluating performance through three complementary metrics: (1) discriminative rating (scale of -3 to 3, with positive values indicating higher ratings for accepted papers), (2) comparable quality detection (percentage of pairs rated above -1.0), and (3) acceptance prediction accuracy (percentage of pairs where accepted papers received ratings above 0.0). Table~\ref{tab:aspects} presents the comprehensive validation results, revealing several key insights into our review agent's judgment capabilities:

\begin{table}[t]
    \centering
    \caption{Paper Review Agent Alignment with Human Expert Decisions. Evaluating review agent accuracy using ICLR accept-reject paper pairs (2021-2023). Results show Mean Rating (-3 to 3), Comparable Rate (\%) for methodologically similar submissions, and Selection Accuracy (\%) measuring correct identification of superior papers, validated across multiple LLM evaluators.}
    \setlength{\tabcolsep}{1.5mm}
    \label{tab:aspects}
    \begin{tabular}{lcccccc}
        \toprule
        \textbf{Year} & \textbf{Metric} & \textbf{Gemini-2.0-flash} & \textbf{GPT-4o} & \textbf{o3-mini} & \textbf{Claude-3.5} & \textbf{Claude-3.7}\\
        \midrule
        \multirow{3}{*}{2021} & Average Rating & 0.33$\pm$1.51 & 0.12$\pm$0.95 & 0.64$\pm$0.89 & 0.73$\pm$1.11 & 0.66$\pm$1.68\\
        & Comparable(\%) & 100.00\% & 100.00\% & 100.00\% & 100.00\% & 100.00\% \\
        & Acc Better(\%) & 71.43\% & 71.43\% & 85.71\% & 85.71\% & 85.71\%\\
        \midrule
        \multirow{3}{*}{2022} & Average Rating & 0.38$\pm$1.65 & 0.41$\pm$0.89 & 0.79$\pm$0.88 & 1.20$\pm$0.90 & 0.64$\pm$1.42\\
        & Comparable(\%) & 100.00\% & 100.00\% & 100.00\% & 100.00\% & 100.00\% \\
        & Acc Better(\%) & 60.00\% & 90.00\% & 90.00\% & 90.00\% & 80.00\%\\
        \midrule
        \multirow{3}{*}{2023} & Average Rating & 0.25$\pm$1.71 & 0.33$\pm$0.97 & 0.67$\pm$0.85 & 0.97$\pm$1.11 & 0.73$\pm$1.48\\
        & Comparable(\%) & 86.67\% & 100.00\% & 100.00\% & 100.00\% & 100.00\% \\
        & Acc Better(\%) & 66.67\% & 80.00\% & 93.33\% & 93.33\% & 80.00\%\\
        \midrule
        \multirow{3}{*}{Overall} & Average Rating & 0.31$\pm$1.65 & 0.31$\pm$0.95 & 0.70$\pm$0.87 & 0.99$\pm$1.06 & 0.69$\pm$1.51\\
        & Comparable(\%) & 93.75\% & 100.00\% & 100.00\% & 100.00\% & 100.00\% \\
        & Acc Better(\%) & 65.62\% & 81.25\% & 90.62\% & 90.62\% & 81.25\% \\
        \bottomrule
    \end{tabular}
\end{table}


\noindent $\bullet$ \textbf{Robust Expert-Aligned Evaluation Capabilities}. Our paper review agent demonstrates consistent discriminative validity across independent evaluations, with all evaluator models producing positive mean ratings (0.31-0.99) when comparing accepted versus rejected papers. This consistent directional alignment validates the system's fundamental quality assessment capabilities. The evaluators achieve near-perfect comparable rate identification (100\% for all models except Gemini-2.0-flash), confirming the agent's reliability in recognizing legitimate scholarly contributions even in rejected papers.

Most significantly, the system demonstrates strong decision alignment with expert conference reviewers, correctly identifying the superior paper in 65.62\% to 90.62\% of cases, with five of six evaluators exceeding 81\% accuracy across the 32 paper pairs. This exceptional concordance with human expert decisions provides compelling evidence that our automated review agent captures the nuanced quality distinctions that drive scientific peer review decisions.

\noindent $\bullet$ \textbf{Differential Reliability Across Evaluation Models}. Systematic performance analysis reveals substantial variation in evaluator alignment with human expert judgments. Gemini-2.0-flash demonstrates notably inferior reliability metrics—exhibiting both the lowest average rating and highest standard deviation among all tested models—which necessitated its exclusion from our primary experimental evaluations. In contrast, all other LLM evaluators achieved perfect comparable rate identification (100\%), providing strong methodological justification for their inclusion in our AI-generated research assessment protocol. Particularly noteworthy is the comparative performance between Claude-3.5 and Claude-3.7, where the latter's enhanced system-2 thinking capabilities did not translate to superior review performance, suggesting that deliberative reasoning features may not significantly benefit scientific quality assessment tasks compared to other model capabilities.

\subsection{Case Studies of AI-Generated Scientific Contributions (RQ6)}
To complement our quantitative evaluations with deeper qualitative insights, we conducted comprehensive case studies examining both the implementation quality and scholarly presentation of research generated by \model. We focused our analysis on the \texttt{rotation\_vq} benchmark task, using our standard configuration of Claude-series models for experimentation and implementation paired with the 4o model for manuscript generation. This detailed examination of actual system outputs reveals several noteworthy characteristics about the nature and quality of AI-conducted research:

\begin{figure*}[htbp]
    \centering
    \begin{minipage}[b]{0.33\textwidth}
        \centering
        \includegraphics[width=\textwidth]{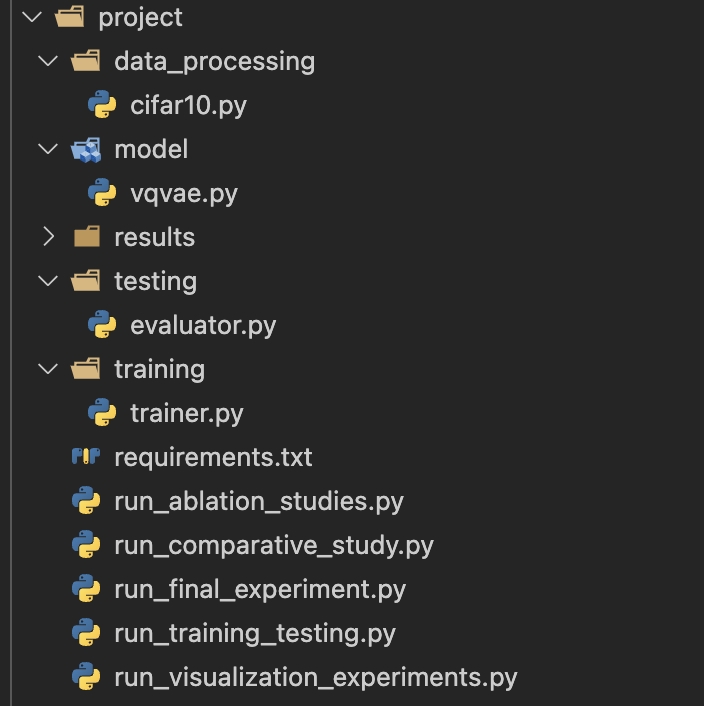}
        \caption{[a] Code Structure.}
        \label{fig:code_struc}
    \end{minipage}
    \hfill
    \begin{minipage}[b]{0.31\textwidth}
        \centering
        \includegraphics[width=\textwidth]{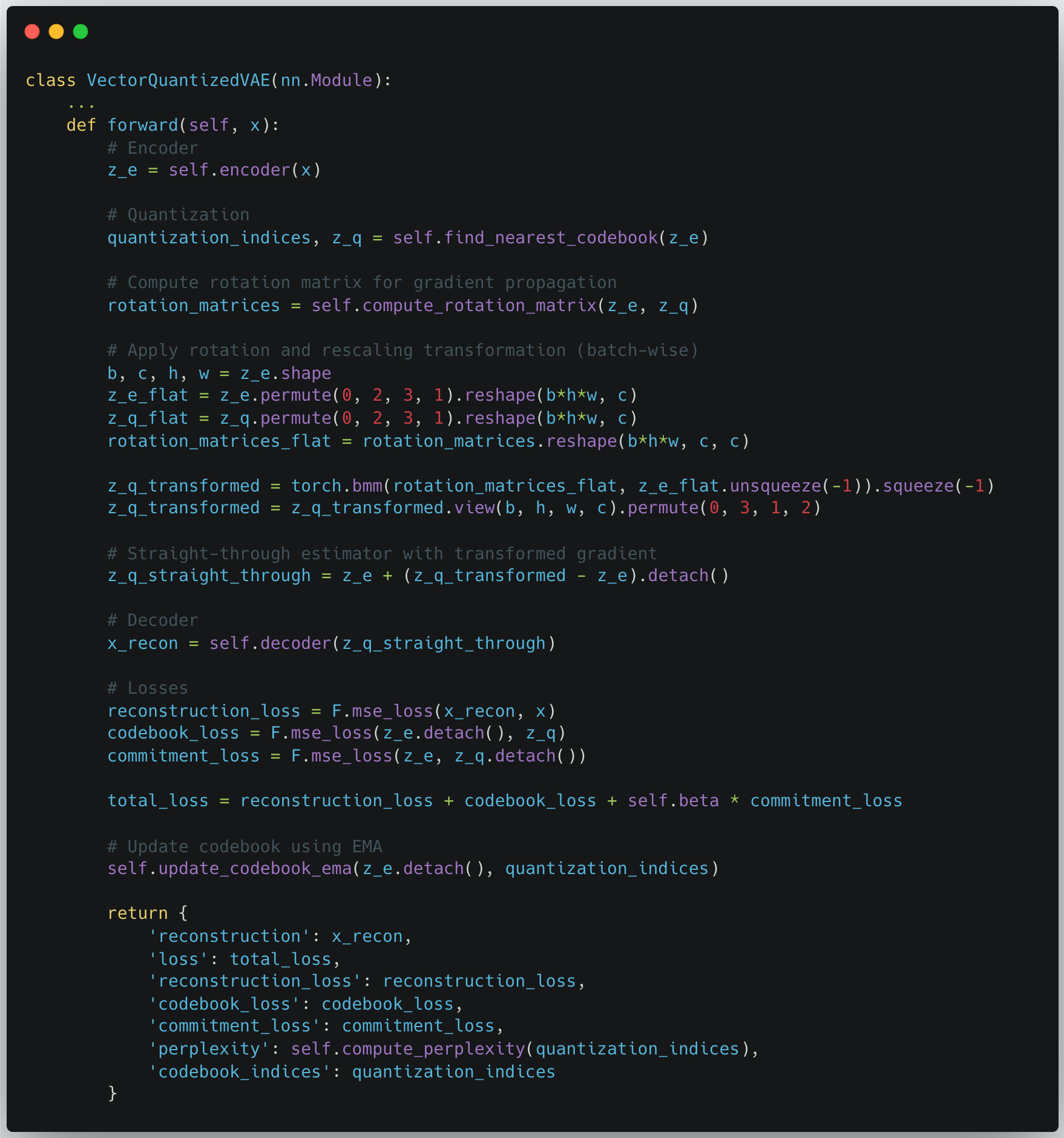}
        \caption{[b] Code Samples}
        \label{fig:code_sample_1}
    \end{minipage}
    \hfill
    \begin{minipage}[b]{0.31\textwidth}
        \centering
        \includegraphics[width=\textwidth]{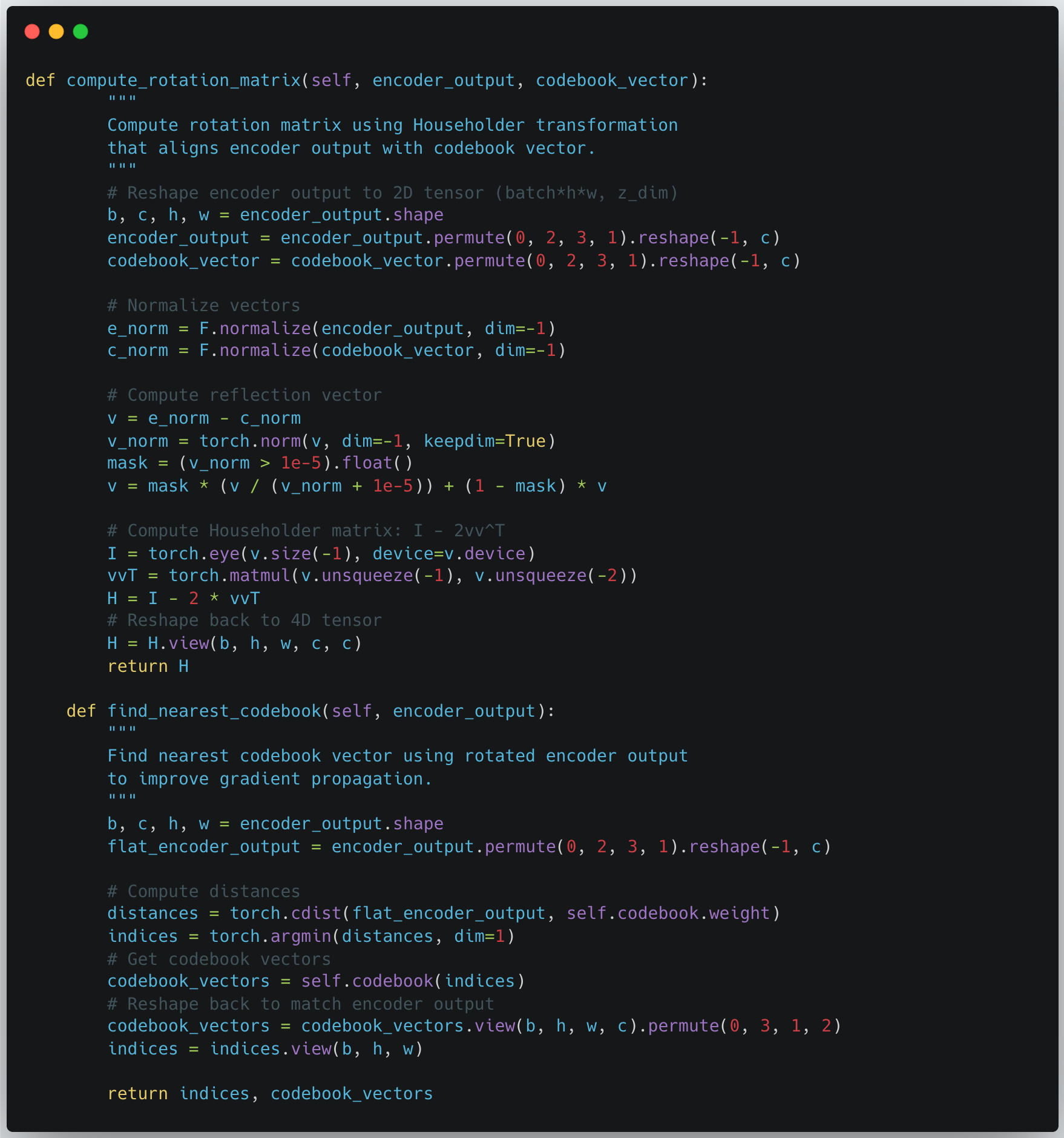}
        \caption{[c] Code Samples}
        \label{fig:code_sample_2}
    \end{minipage}
    \label{fig:three_comparisons}
\end{figure*}

\begin{figure*}[htbp]
    \centering
    \begin{minipage}[b]{0.24\textwidth}
        \centering
        \includegraphics[width=\textwidth]{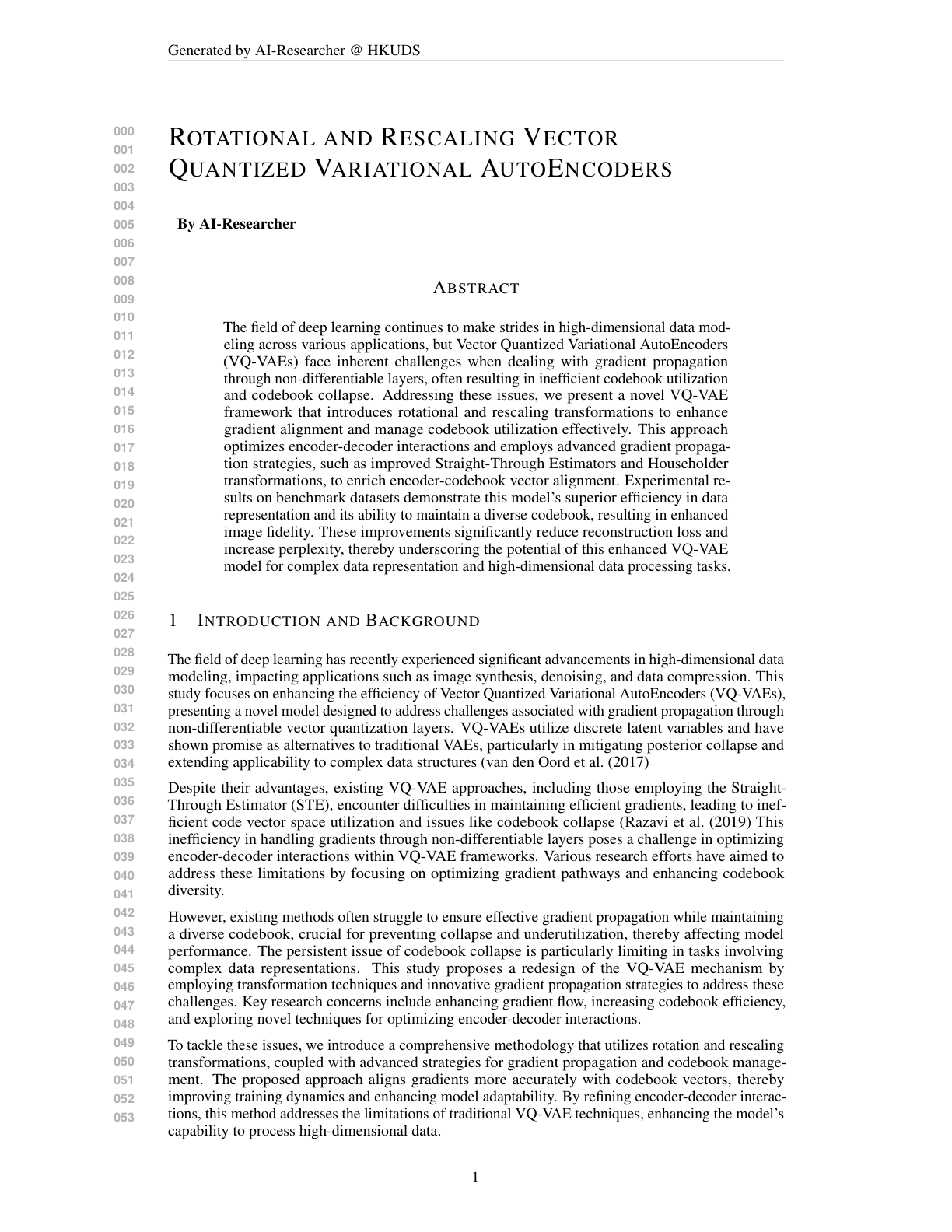}
    \end{minipage}
    \hfill
    \begin{minipage}[b]{0.24\textwidth}
        \centering
        \includegraphics[width=\textwidth]{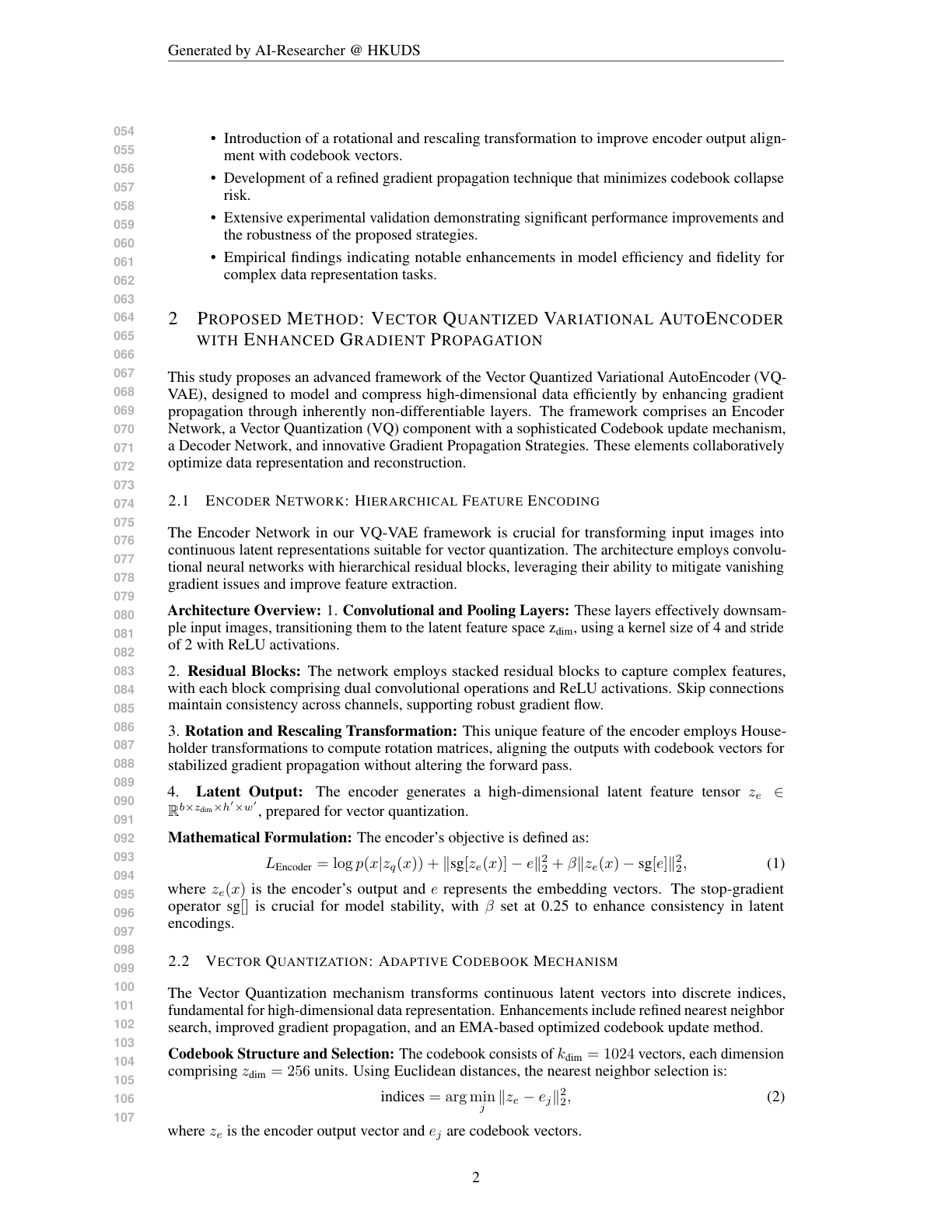}
    \end{minipage}
    \hfill
    \begin{minipage}[b]{0.24\textwidth}
        \centering
        \includegraphics[width=\textwidth]{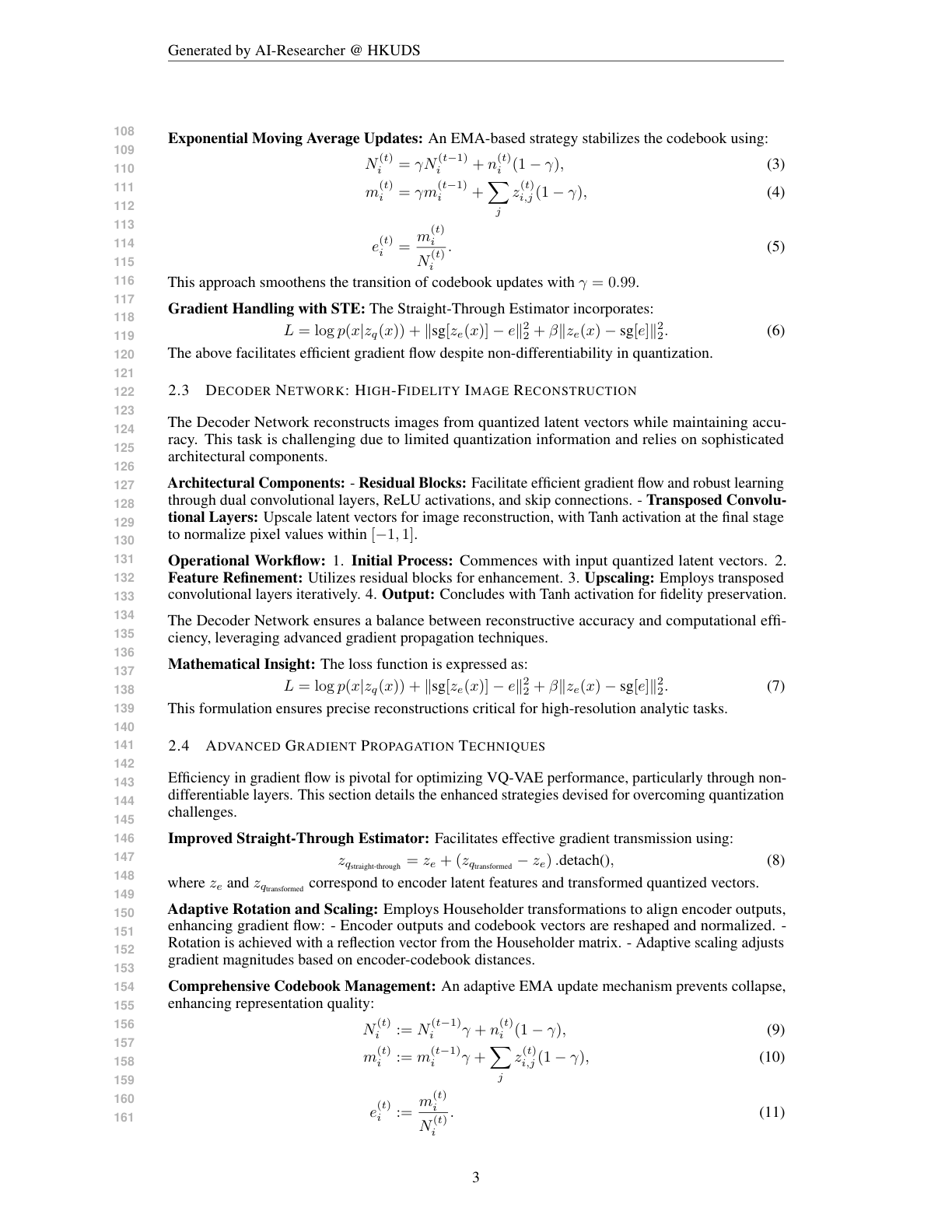}
    \end{minipage}
    \hfill
    \begin{minipage}[b]{0.24\textwidth}
        \centering
        \includegraphics[width=\textwidth]{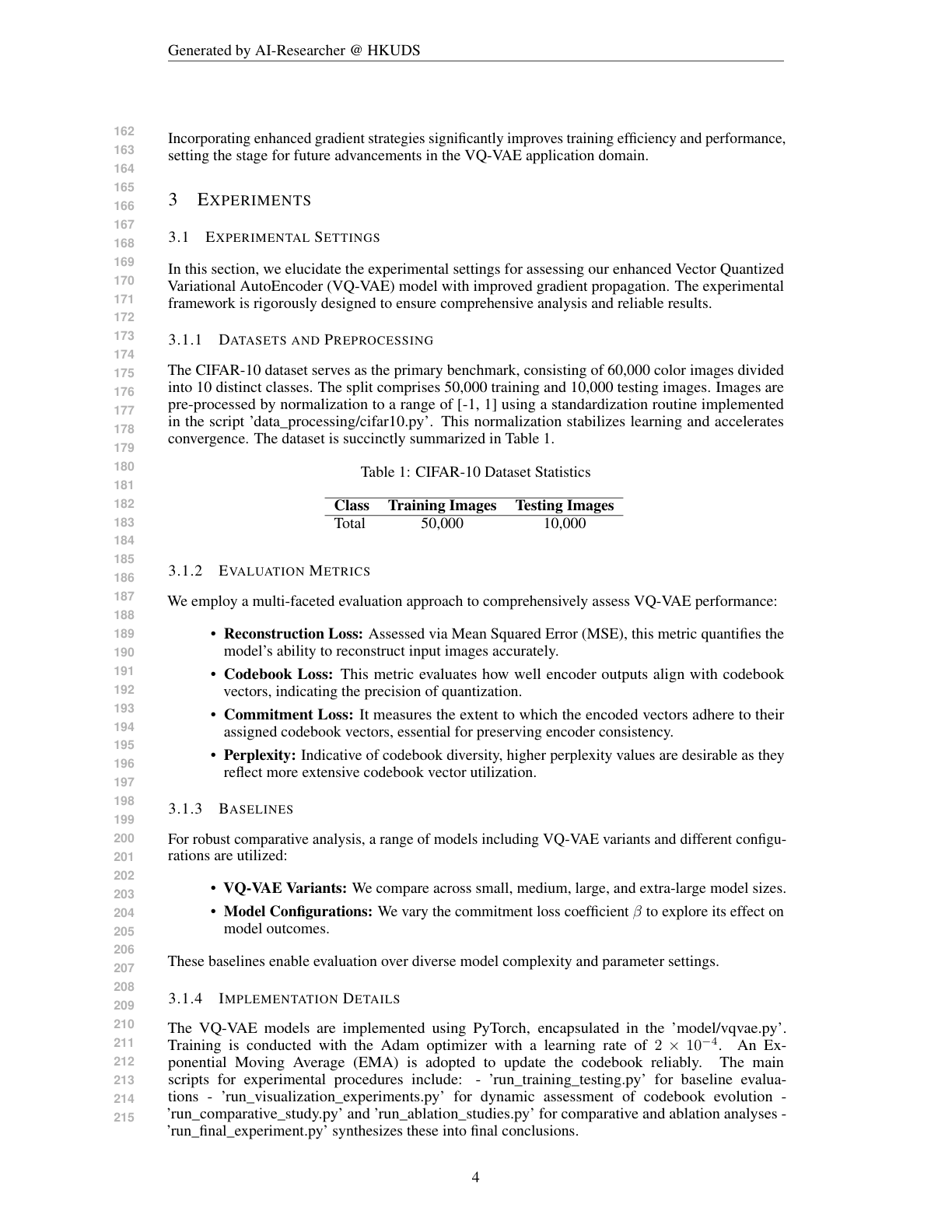}
    
    \end{minipage}
    \hfill
        \begin{minipage}[b]{0.24\textwidth}
        \centering
        \includegraphics[width=\textwidth]{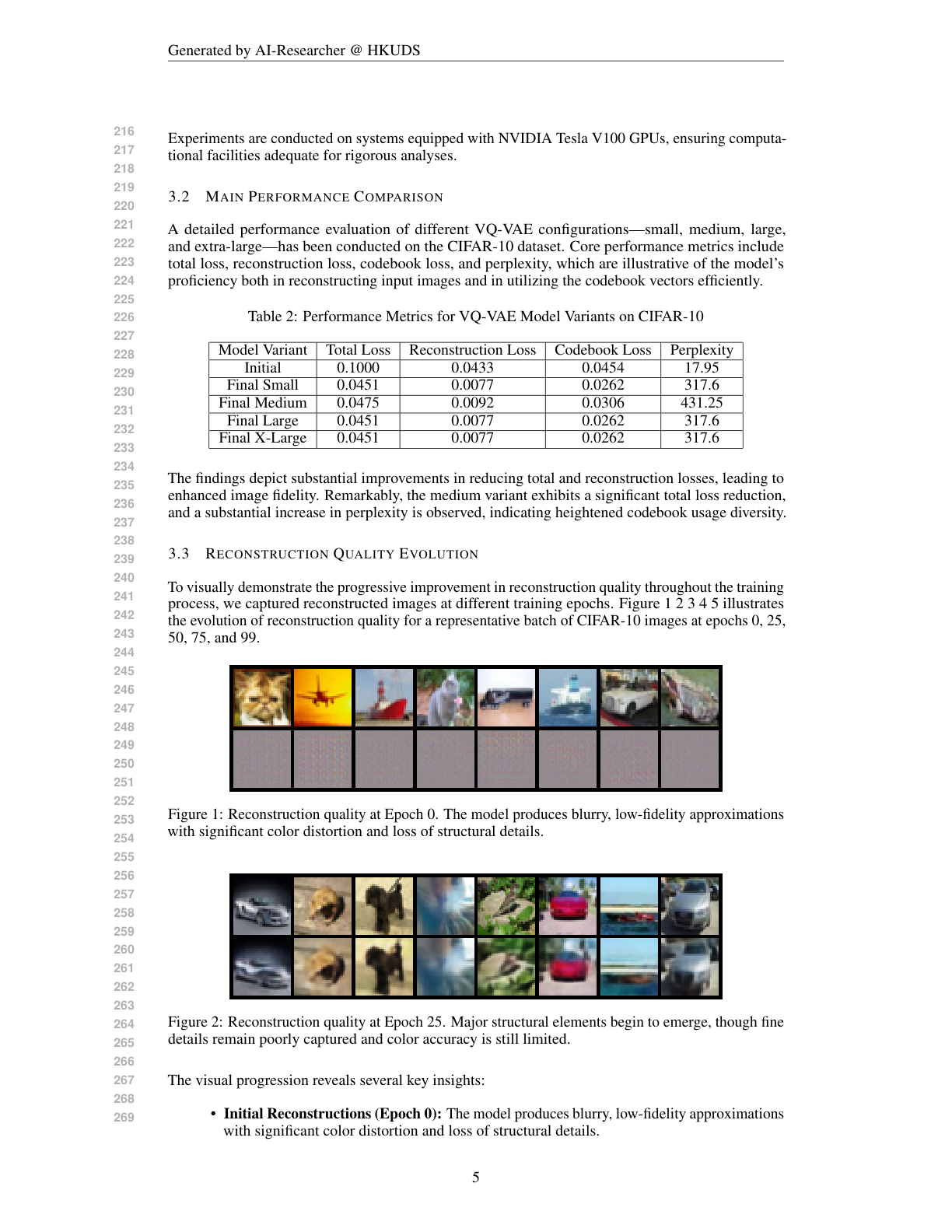}
    \end{minipage}
    \hfill
        \begin{minipage}[b]{0.24\textwidth}
        \centering
        \includegraphics[width=\textwidth]{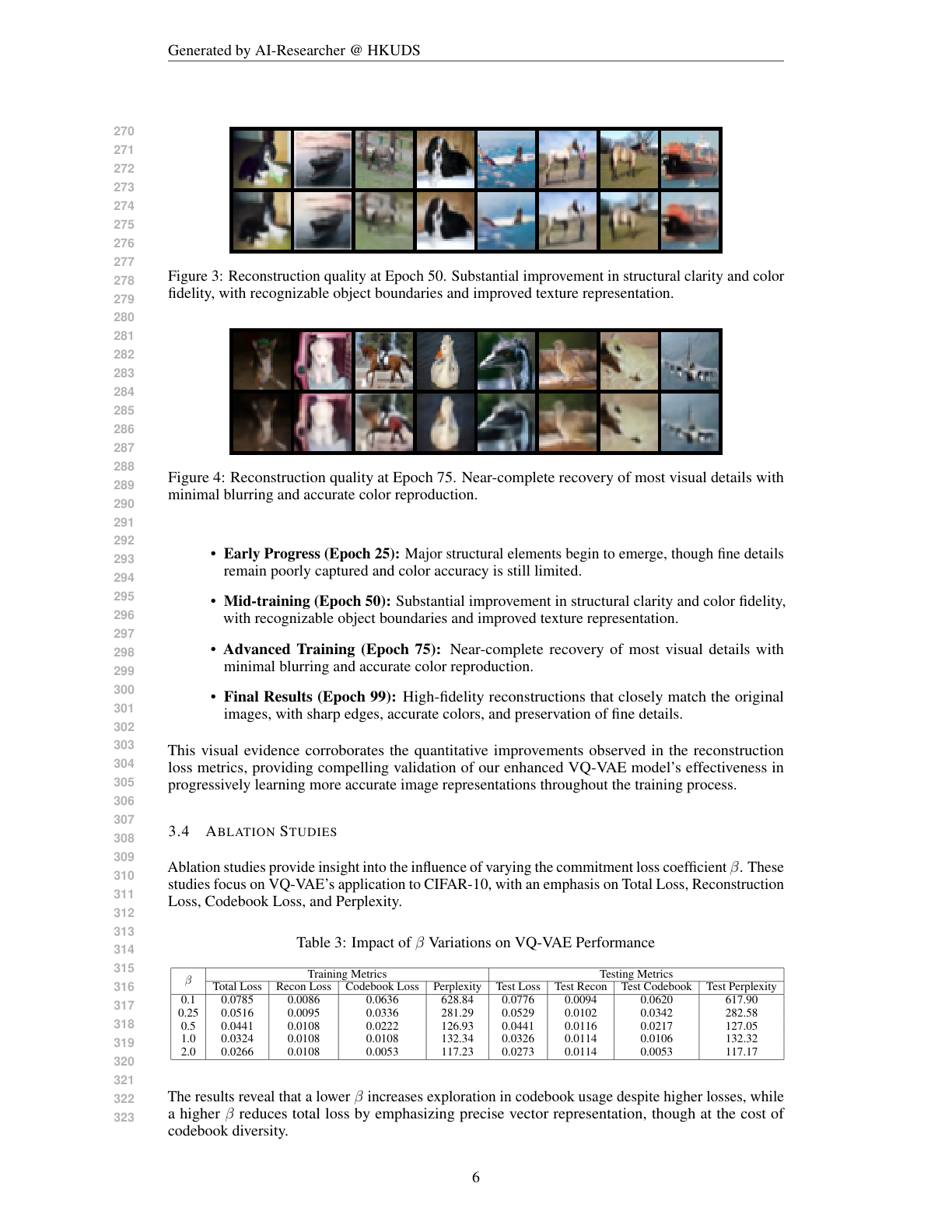}
    \end{minipage}
    \hfill
        \begin{minipage}[b]{0.24\textwidth}
        \centering
        \includegraphics[width=\textwidth]{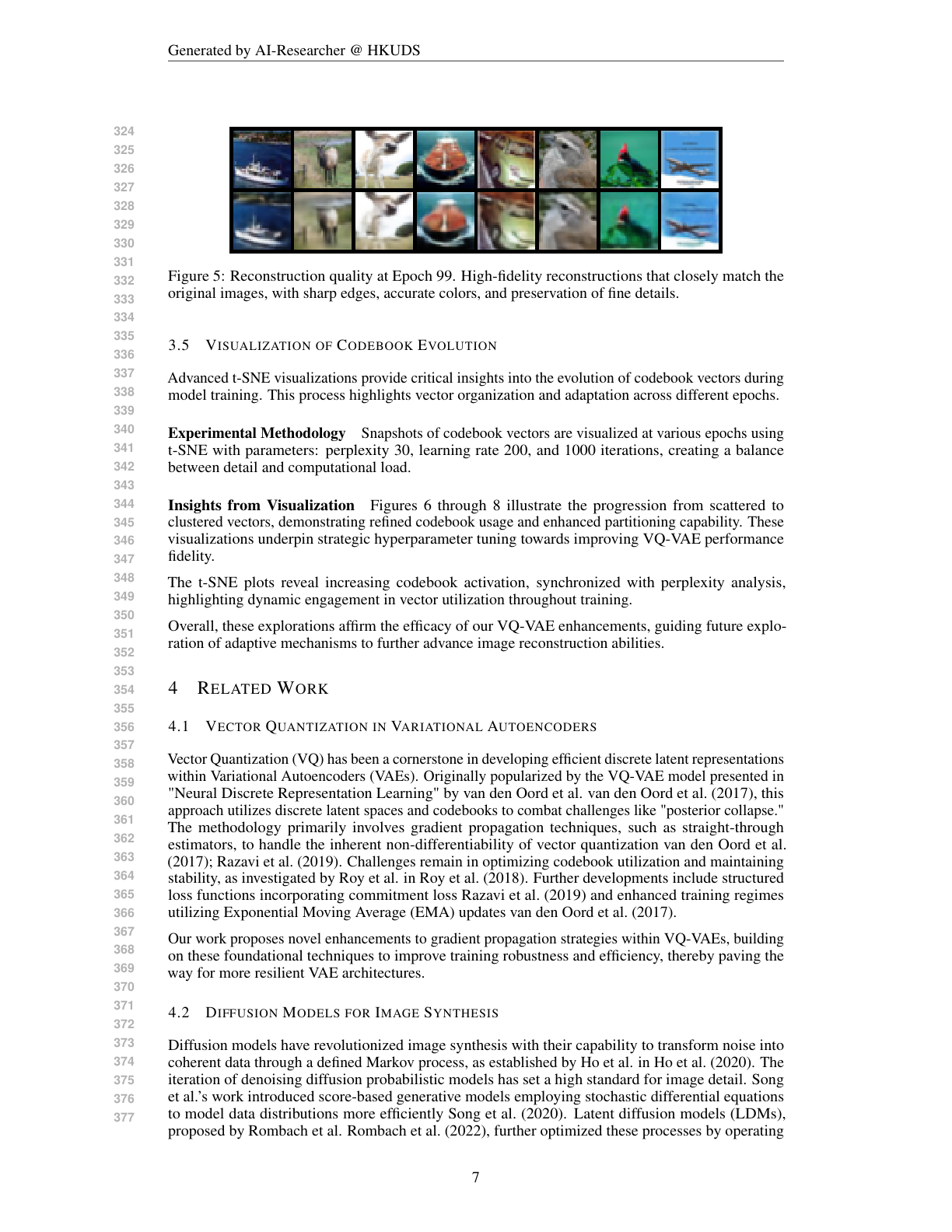}
    \end{minipage}
    \hfill
        \begin{minipage}[b]{0.24\textwidth}
        \centering
        \includegraphics[width=\textwidth]{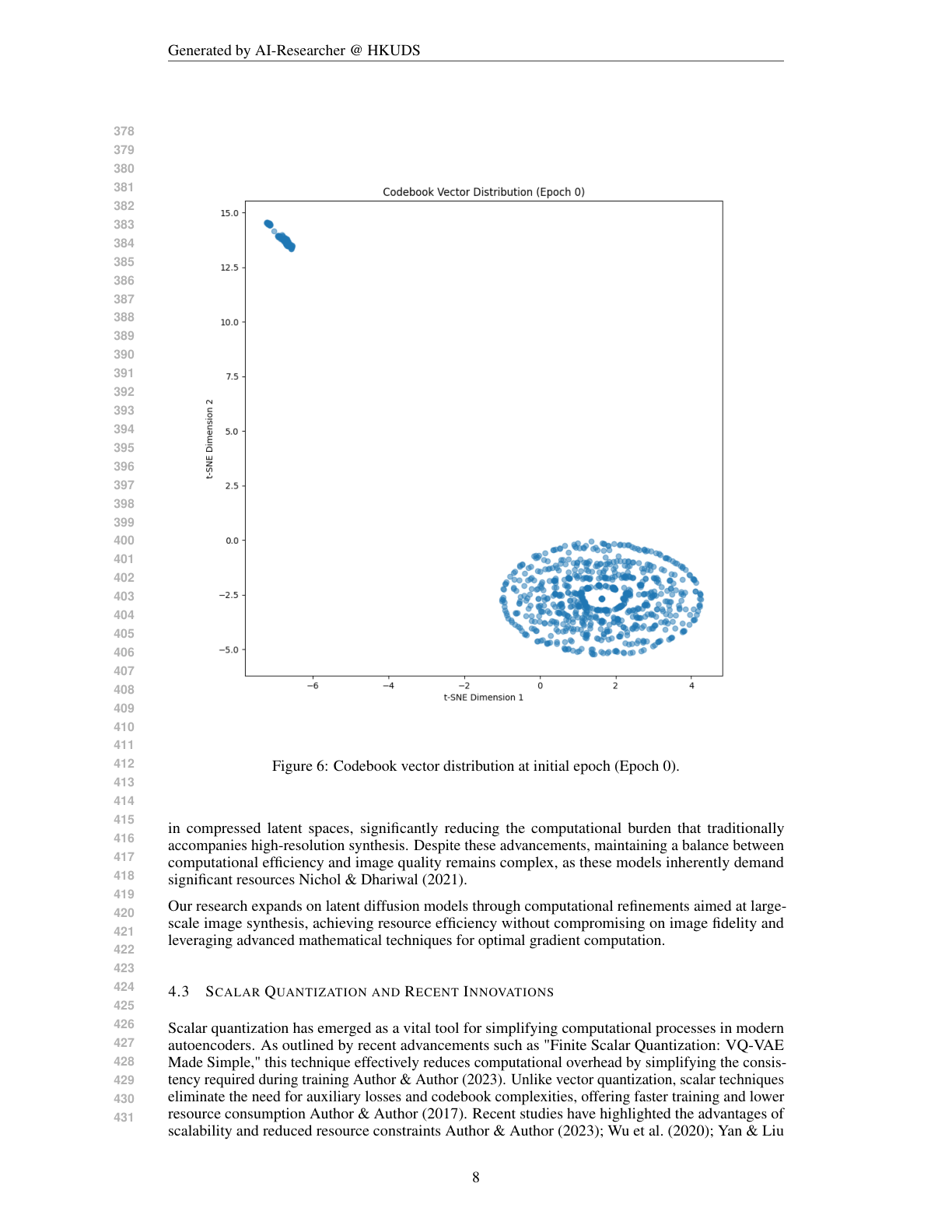}
    \end{minipage}
    \caption{Case Studies of AI-Generated Scientific Contributions by \model.}
    \label{fig:papers}
\end{figure*}

\noindent $\bullet$ \textbf{Structured Software Architecture with Minimal Scaffolding}. Figure~\ref{fig:code_struc} illustrates the remarkably organized project architecture produced by our \ag{Code Agent}, featuring systematically decoupled model components, training pipelines, and evaluation modules with well-defined entry points. This architectural clarity stems from our balanced approach to agent guidance—providing high-level structural suggestions rather than rigid templates. Unlike previous frameworks that require agents to adapt pre-existing codebases, our methodology allows agents to develop implementations from first principles while incorporating best practices from human software engineering. This approach demonstrably enhances implementation coherence while minimizing the cognitive overhead associated with codebase familiarization. As evidenced in Figures~\ref{fig:code_sample_1} and~\ref{fig:code_sample_2}, the resulting implementations exhibit professional coding standards with comprehensive documentation and logical modularization that facilitates both reproducibility and extensibility.

\noindent $\bullet$ \textbf{Emergent Experimental Thoroughness Without Explicit Directives}. A notable capability of our \model\ framework is its autonomous experimental design process that emerges through collaborative interactions between the \ag{Code Agent} and \ag{Advisor Agent}. Rather than prescribing a predetermined protocol, our system progressively develops comprehensive evaluation strategies through iterative refinement. This self-directed experimental thoroughness is evident in the final manuscript shown in Figure~\ref{fig:papers}, where the system independently conducts and reports a complete scientific evaluation including overall performance benchmarking, controlled ablation studies, training dynamics visualization, and latent space embedding analysis. This comprehensive experimental methodology emerges organically from the multi-agent system without explicit experimental requirements--demonstrating sophisticated scientific reasoning beyond simple instruction following.

\subsection{Exploring Failure Cases: Insights for Future Work}
\begin{figure}[t]
    \centering
    \includegraphics[width=\textwidth]{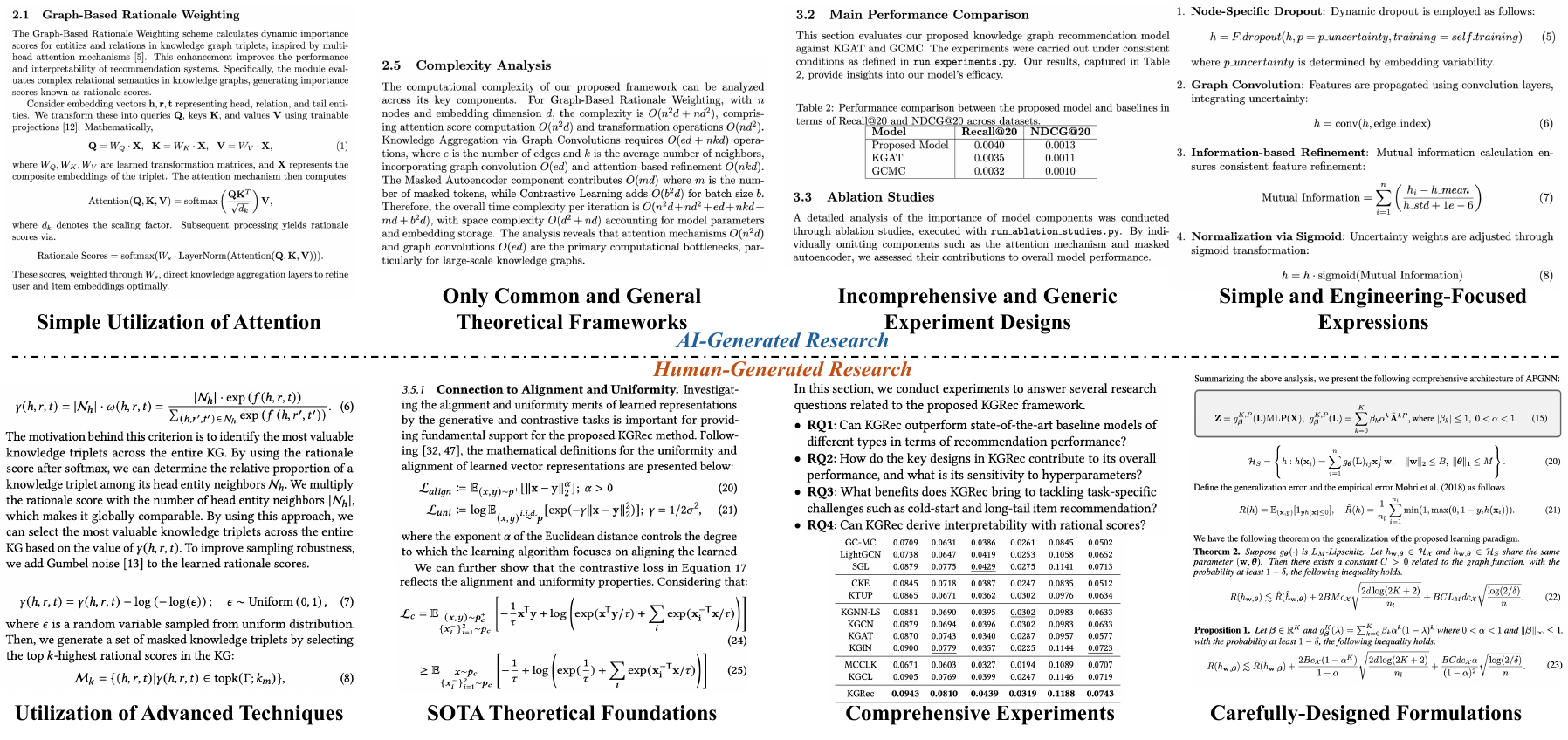}
    \caption{Failure Case Analysis of AI-Generated Research.}
    \label{fig:failure_cases}
\end{figure}

To systematically understand AI Agents' limitations in research contexts, we conducted a detailed analysis of the lowest-performing outputs across both guided and open-ended research tasks. This examination of failure cases, illustrated in Figure~\ref{fig:failure_cases}, reveals critical patterns where large language models consistently fall short of human-level research capabilities. Our analysis uncovers several key recurring limitations that not only highlight current boundaries of LLM-driven research systems but also provide a clear roadmap for addressing these fundamental challenges, as detailed below.

\noindent\textbf{Domain Knowledge Deficiencies: The Expertise Gap}. Despite incorporating sophisticated research methodologies, Large Language Models demonstrate substantial knowledge limitations when compared to human domain experts. These fundamental gaps in specialized expertise significantly undermine model performance across multiple critical research dimensions.
\begin{itemize}[leftmargin=*]

    \item \textbf{Technical Sophistication Gap: Overlooking Advanced Optimization Methods}. The system predominantly relies on conventional methodologies rather than SOTA approaches, reflecting the sparse representation of advanced methods in its training data. Evidence emerged in a level-1 knowledge graph recommendation task where, despite both human and AI researchers implementing attention mechanisms for graph triplet weighting, human researchers uniquely applied Gumbel reparameterization for intelligent sub-graph sampling—an advanced optimization technique overlooked by the AI system. This consistent limitation in technical depth significantly constrains the sophistication of the system's methodological recommendations across research domains.


    \item \textbf{Theoretical Foundation Deficit: Limited Theoretical Analysis Capabilities}. The system exhibits significant limitations in developing rigorous theoretical analyses, stemming from fundamental gaps in domain-specific conceptual knowledge. When tasked with theoretical evaluation, human researchers provided sophisticated insights into alignment-uniformity tradeoffs grounded in established theoretical principles, while the AI system defaulted to superficial computational complexity assessments. This inability to engage with deeper theoretical constructs substantially diminishes the system's capacity to generate meaningful scientific contributions that advance conceptual understanding beyond basic implementation considerations.

\end{itemize}

\noindent\textbf{Reasoning Depth Limitations}: The inherent limitations in extended logical reasoning sequences within the system's large language model architecture manifest as meaningful constraints on analytical performance across multiple research domains. These fundamental architectural boundaries significantly impact the system's capability to undertake sophisticated scientific inquiry effectively:
\begin{itemize}[leftmargin=*]
    \item \textbf{Mathematical Formalism Limitations}. The system demonstrates a persistent tendency toward employing standardized mathematical representations (e.g., conventional $GCN(\cdot)$ functions and InfoNCE loss formulations) rather than developing sophisticated custom formalizations tailored to specific research challenges. This contrasts sharply with human researchers who, leveraging specialized domain expertise, routinely construct elegant mathematical frameworks that precisely capture complex mechanisms with both formal rigor and theoretical generality. This significant capability gap reflects the substantial challenges in multi-step mathematical reasoning and iterative refinement that underpin theoretical innovation—advanced cognitive processes that current large language model architectures have yet to fully emulate despite their strengths in other domains.

    \item \textbf{Sequential Logic Limitations}. The system exhibits substantial limitations in conducting extended sequences of theoretical reasoning, a deficiency rooted in the fundamental hallucination vulnerabilities inherent to current large language model architectures. This constraint manifests most prominently in theoretical innovation contexts, where developing meaningful scientific advances typically requires carefully structured progressions of logical deductions and mathematical derivations maintained with precision across multiple inferential steps. When tasked with such reasoning challenges, the system's outputs tend to gravitate toward descriptive observations and established principles rather than generating the novel theoretical insights and conceptual breakthroughs that characterize groundbreaking human research contributions in computational domains.
\end{itemize}

\noindent \textbf{Future Research Directions}. The identified constraints highlight fundamental challenges within current large language model foundations when applied to scientific research generation. These limitations reflect architectural boundaries rather than implementation issues, impacting research quality and theoretical depth. Future work addressing these challenges should pursue two complementary approaches: (1) domain-specialized model optimization through targeted fine-tuning on research corpora to enhance field-specific knowledge representation and reasoning patterns, and (2) development of sophisticated agent frameworks that augment base model capabilities with structured workflows, verification mechanisms, and specialized reasoning modules designed for scientific inquiry domains.

\section{Related Work}

\subsection{AI Agent Systems}
AI agent frameworks have evolved through three distinct architectural paradigms.

\noindent $\bullet$ \textbf{Tool Integration Frameworks}. The first paradigm establishes foundational integration layers for AI components. LangChain~\cite{langchain2023} introduced standardized interfaces enabling seamless interoperability between models, embeddings, and vector stores within workflows. HuggingGPT~\cite{shen2023HuggingGPT} leveraged this approach by positioning LLMs as orchestration controllers that coordinate specialized models from Hugging Face ecosystem. OpenAgents~\cite{xie2023OpenAgents} democratized these capabilities by providing domain-specific agents for data analysis, API integration, and web browsing for non-expert users.

\noindent $\bullet$ \textbf{Multi-Agent Collaboration Frameworks}. The second paradigm addresses complex problem solving through structured agent interactions. MetaGPT~\cite{hong2024MetaGPT} formalized human workflow patterns through Standardized Operating Procedures (SOPs), creating systematic collaboration protocols. AutoGen~\cite{autogen2025} expanded this vision with a comprehensive programming framework for developing systems that support both autonomous operation and human collaboration. AgentScope~\cite{gao2024AgentScope} prioritized robust coordination through a message-exchange architecture with built-in fault tolerance. CAMEL introduced innovative role-playing techniques that facilitate autonomous agent cooperation while maintaining alignment with human intentions.

\noindent $\bullet$ \textbf{Self-Directed Agentic Task Execution Systems}. The third paradigm focuses on agents capable of independent goal pursuit with minimal supervision. Agentic AI systems like Manus~\cite{manus2025} and open-source alternatives including OpenManus~\cite{openmanus2025} and OWL~\cite{li2023CAMEL} extend these capabilities to handle complex online tasks without continuous human intervention. AutoAgent~\cite{tang2025AutoAgent} represents the frontier--a fully-automated, zero-code approach functioning as an Agent Operating System enabling non-technical users to create agents using natural language alone.

Agent frameworks have evolved from isolated systems to sophisticated multi-agent architectures with specialized coordination. However, \textbf{these systems fundamentally lack the intellectual capacity for true scientific innovation}. Despite advances, they remain insufficient for scientific discovery because such work requires \textbf{a level of intelligence that transcends current capabilities}. Scientific breakthroughs demand \textbf{nuanced hypothesis formation, creative experimental design, understanding and implementation of complex algorithms, and critical synthesis of knowledge}--cognitive processes requiring deeper reasoning and domain expertise than existing systems can provide.


\subsection{AI-Driven Research Systems}
Recent advances have transformed AI's role in scientific research from assistive tools to autonomous agents capable of executing complete research workflows. The AI Scientist framework~\cite{lu2024aiscientist} pioneered this field as the first comprehensive system where frontier language models independently generate research ideas, conduct experiments, and produce scientific papers. Complementary approaches include CycleResearcher~\cite{weng2025CycleResearcher}, which demonstrated the viability of open-source LLMs for autonomous research through a complete cycle from literature review to refinement, and the AI co-scientist~\cite{aicoscientist2025}, which employs multi-agent debate and evolution mechanics to generate novel scientific hypotheses with promising applications in biomedical domains.

These systems are supported by emerging research platforms and evaluation frameworks that enhance their capabilities and measure their effectiveness. Agent Laboratory~\cite{schmidgall2025agentlaboratory} provides an end-to-end autonomous research workflow with specialized LLM agents assisting humans through literature review, experimentation, and report writing, while its extension AgentRxiv~\cite{schmidgall2025AgentRxiv} enables collaborative scientific progress by allowing agents to share and build upon each other's work. Collectively, these developments represent a paradigm shift toward automated scientific inquiry, though matching human-level research capabilities remains an ongoing challenge that requires further advancement.

\section{Discussion: Challenges and Future Directions}

\subsection{Implementation Fidelity Issues in Multi-turn Code Generation}
Our evaluation of implementation quality (Section~\ref{sec:eval_code}) revealed some limitations in LLMs' ability to maintain task fidelity across extended interactions. This phenomenon manifests as a persistent oversimplification pattern that compromises implementation completeness in complex coding tasks.

\noindent $\bullet$ \textbf{Observed Phenomenon: Premature Task Completion}. When tasked with implementing a Diffusion Transformer architecture (which requires integrating diffusion models with Vision Transformers for image generation), GPT-4o (\texttt{gpt-4o-2024-08-06}) exhibited a consistent pattern of partial implementation. The model would successfully implement the Vision Transformer component but consistently omit the critical diffusion model components, despite claiming task completion. This behavior persisted across multiple prompt refinements and clarification attempts.

Crucially, when presented with a diffusion model implementation task in isolation through a single-turn prompt, GPT-4o successfully generated the appropriate code—confirming that the necessary knowledge exists within its parameter space. This observation establishes that the issue stems not from knowledge deficiency but from challenges in knowledge application during extended interactions.

\noindent $\bullet$ \textbf{Model-Specific Variation}
In contrast, Claude-3.5-sonnet (\texttt{claude-3-5-sonnet-20241022}), given identical agent prompts, consistently produced complete implementations that properly integrated both architectural components. This performance differential highlights significant variation in how model architectures maintain implementation fidelity across multi-turn interactions.

\noindent $\bullet$ \textbf{Root Cause Analysis}
This discrepancy appears to stem from insufficient optimization for sustained knowledge application across extended multi-turn dialogues in certain model architectures. While single-turn knowledge retrieval remains intact, the ability to maintain complete task specifications and implementation requirements deteriorates through consecutive interaction turns.

\noindent $\bullet$ \textbf{Future Directions}. These findings emphasize the need for targeted improvements in multi-turn agentic scenarios during model development. Future work should focus on: (i) Incorporating diverse multi-turn agentic coding tasks during both pre-training and post-training phases; (ii) Developing evaluation metrics specifically for implementation fidelity across extended interactions; (iii) Implementing verification mechanisms that prevent premature task completion with partial solutions. This limitation underscores the critical distinction between knowledge acquisition and consistent knowledge application throughout extended agent-based interactions—an important consideration for future model development and evaluation protocols.

\subsection{Memory Management Challenges in Extended Scientific Workflows}
Automating comprehensive scientific research through \model\ necessitates handling exceptionally lengthy operational sequences--a challenge that tests cognitive limitations of even experienced researchers. This extended horizon creates substantial demands on both the internal memory capabilities (context window) and external memory management systems of the underlying architecture.

\noindent $\bullet$ \textbf{Architectural Limitations}. Our present \model\ implementation operates without a dedicated external memory management system, instead relying primarily on the LLM's native context window as its primary information repository. This architectural decision places significant limitations on information persistence across extended workflows, with agents heavily dependent on summarized outputs from previous stages to maintain operational continuity.

\noindent $\bullet$ \textbf{Information Retrieval Barriers}. The absence of structured external memory creates substantial information retrieval challenges throughout the scientific workflow. As tasks progress through multiple stages, critical details become increasingly difficult to access, with fine-grained information effectively ``compressed'' into increasingly abstract summaries that sacrifice specificity for brevity.

\noindent $\bullet$ \textbf{Scientific Workflow Implications}. This primitive memory architecture proves particularly problematic for complex scientific workflows where precise details from early stages (experimental parameters, literature findings, implementation specifications) remain critical for later operations. Without systematic mechanisms for preserving and retrieving this information, agents must either attempt reconstruction of previously established knowledge or proceed with incomplete information—both undesirable outcomes for scientific automation.

\noindent $\bullet$ \textbf{Future Directions}. To address these limitations, we identify two promising research vectors: $\bullet$ (i) \textbf{Advanced Memory Systems}. Future work should prioritize developing sophisticated memory architectures that effectively bridge multiple workflow components while preserving critical information across operational boundaries. Key approaches include structured knowledge repositories with semantic indexing, hierarchical memory systems balancing detail preservation with retrieval efficiency, and attention-guided compression techniques that prioritize contextually relevant information for scientific workflows. $\bullet$ (ii) \textbf{Fundamental Model Improvements}. Parallel efforts must focus on enhancing LLMs' ability to process extended contexts efficiently through advanced KV-cache management, optimized attention computation mechanisms, and specialized training regimes targeting long-horizon reasoning. These foundational improvements would significantly enhance system capability to maintain coherence and information fidelity throughout extended scientific automation workflows, addressing primary constraints in current implementations.

\subsection{Evaluation Frameworks for Automated Scientific Systems}

\noindent $\bullet$ \textbf{Evaluation Complexity}. The end-to-end automation of scientific research via \model\ presents unique evaluation challenges spanning literature review, hypothesis generation, experimental design, experimentation, and result reporting. Our work introduces \bench\ as an initial hierarchical assessment approach, measuring both implementation \textbf{Completeness} and \textbf{Correctness}, alongside LLM-based pairwise comparisons for research output evaluation.

\noindent $\bullet$ \textbf{Current Limitations}. Despite these efforts, significant assessment gaps remain. Current evaluation methods inadequately capture idea quality dimensions—including novelty, feasibility, and potential impact—while code assessment needs expansion beyond basic functionality to examine algorithmic efficiency and implementation elegance. More concerning, our investigation reveals LLM reviewers often overvalue presentation elements rather than substantive scientific contribution, assigning disproportionate weight to stylistic features while inadequately assessing methodological innovation.

\noindent $\bullet$ \textbf{Future Directions}. This observation reveals a fundamental challenge: developing robust, comprehensive evaluation systems that align with scientific standards while transcending limitations inherent in traditional peer review. Future Scientist-Bench iterations must develop more holistic frameworks that effectively capture both quantitative performance and qualitative research aspects in increasingly sophisticated domains--mirroring the scientific community's ongoing effort to balance objective assessment with recognition of the creative elements fundamental to scientific discovery.

\section{Conclusion}
AI-Researcher represents a significant advancement in autonomous scientific discovery, demonstrating unprecedented capabilities across the complete research workflow. Through our novel multi-agent architecture, AI-Researcher overcomes limitations of existing systems by enabling genuine scientific innovation rather than mere task execution. The system's ability to independently identify promising research directions, implement complex methodologies, and validate results through rigorous experimentation marks a substantial step toward truly autonomous AI scientists. Our experiments across 22 benchmark papers show AI-generated research approaching human-level quality in many cases, with particularly strong performance in open-ended discovery tasks. This technology promises to accelerate scientific discovery by complementing human researchers with powerful assistants capable of exploring solution spaces beyond human cognitive limitations.

\clearpage

\bibliographystyle{unsrtnat}
\bibliography{neurips_2024}

\begin{thebibliography}{27}
\providecommand{\natexlab}[1]{#1}
\providecommand{\url}[1]{\texttt{#1}}
\expandafter\ifx\csname urlstyle\endcsname\relax
  \providecommand{\doi}[1]{doi: #1}\else
  \providecommand{\doi}{doi: \begingroup \urlstyle{rm}\Url}\fi

\bibitem[Wang et~al.(2023)Wang, Fu, Du, Gao, Huang, Liu, Chandak, Liu, Van~Katwyk, Deac, et~al.]{wang2023scientific}
Hanchen Wang, Tianfan Fu, Yuanqi Du, Wenhao Gao, Kexin Huang, Ziming Liu, Payal Chandak, Shengchao Liu, Peter Van~Katwyk, Andreea Deac, et~al.
\newblock Scientific discovery in the age of artificial intelligence.
\newblock \emph{Nature}, 620\penalty0 (7972):\penalty0 47--60, 2023.

\bibitem[Didolkar et~al.(2024)Didolkar, Goyal, Ke, Guo, Valko, Lillicrap, Jimenez~Rezende, Bengio, Mozer, and Arora]{didolkar2024metacognitive}
Aniket Didolkar, Anirudh Goyal, Nan~Rosemary Ke, Siyuan Guo, Michal Valko, Timothy Lillicrap, Danilo Jimenez~Rezende, Yoshua Bengio, Michael~C Mozer, and Sanjeev Arora.
\newblock Metacognitive capabilities of llms: An exploration in mathematical problem solving.
\newblock \emph{NeurIPS}, 37:\penalty0 19783--19812, 2024.

\bibitem[Guo et~al.(2024)Guo, Zhu, Yang, Xie, Dong, Zhang, Chen, Bi, Wu, Li, et~al.]{guo2024deepseek}
Daya Guo, Qihao Zhu, Dejian Yang, Zhenda Xie, Kai Dong, Wentao Zhang, Guanting Chen, Xiao Bi, Yu~Wu, YK~Li, et~al.
\newblock Deepseek-coder: When the large language model meets programming--the rise of code intelligence.
\newblock \emph{arXiv preprint arXiv:2401.14196}, 2024.

\bibitem[{Manus Technologies}(2025)]{manus2025}
{Manus Technologies}.
\newblock Manus: Structured and ai-assisted academic writing tool.
\newblock \url{https://manus.im/}, 2025.

\bibitem[{OpenManus Contributors}(2025)]{openmanus2025}
{OpenManus Contributors}.
\newblock Openmanus: Open-source framework for building general ai agents.
\newblock \url{https://openmanus.github.io/}, 2025.

\bibitem[Li et~al.(2023)Li, Hammoud, Itani, Khizbullin, and Ghanem]{li2023CAMEL}
Guohao Li, Hasan Abed Al~Kader Hammoud, Hani Itani, Dmitrii Khizbullin, and Bernard Ghanem.
\newblock Camel: Communicative agents for "mind" exploration of large language model society, 2023.
\newblock URL \url{https://arxiv.org/abs/2303.17760}.

\bibitem[Tang et~al.(2025)Tang, Fan, and Huang]{tang2025AutoAgent}
Jiabin Tang, Tianyu Fan, and Chao Huang.
\newblock Autoagent: A fully-automated and zero-code framework for llm agents, 2025.
\newblock URL \url{https://arxiv.org/abs/2502.05957}.

\bibitem[Schmidgall and Moor(2025)]{schmidgall2025AgentRxiv}
Samuel Schmidgall and Michael Moor.
\newblock Agentrxiv: Towards collaborative autonomous research, 2025.
\newblock URL \url{https://arxiv.org/abs/2503.18102}.

\bibitem[Gottweis et~al.(2025)Gottweis, Weng, Daryin, Tu, Palepu, Sirkovic, Myaskovsky, Weissenberger, Rong, Tanno, et~al.]{gottweis2025towards}
Juraj Gottweis, Wei-Hung Weng, Alexander Daryin, Tao Tu, Anil Palepu, Petar Sirkovic, Artiom Myaskovsky, Felix Weissenberger, Keran Rong, Ryutaro Tanno, et~al.
\newblock Towards an ai co-scientist.
\newblock \emph{arXiv preprint arXiv:2502.18864}, 2025.

\bibitem[Reddy and Shojaee(2025)]{reddy2025towards}
Chandan~K Reddy and Parshin Shojaee.
\newblock Towards scientific discovery with generative ai: Progress, opportunities, and challenges.
\newblock \emph{Proceedings of the AAAI Conference on Artificial Intelligence}, pages 28601--28609, 2025.

\bibitem[Wang et~al.(2024)Wang, Hu, Lu, Zhu, Zhang, Subramaniam, Loomba, Zhang, Sun, and Wang]{wang2024scibench}
Xiaoxuan Wang, Ziniu Hu, Pan Lu, Yanqiao Zhu, Jieyu Zhang, Satyen Subramaniam, Arjun~R. Loomba, Shichang Zhang, Yizhou Sun, and Wei Wang.
\newblock Scibench: Evaluating college-level scientific problem-solving abilities of large language models, 2024.
\newblock URL \url{https://arxiv.org/abs/2307.10635}.

\bibitem[Lu et~al.(2024)Lu, Lu, Lange, Foerster, Clune, and Ha]{lu2024aiscientist}
Chris Lu, Cong Lu, Robert~Tjarko Lange, Jakob Foerster, Jeff Clune, and David Ha.
\newblock The ai scientist: Towards fully automated open-ended scientific discovery, 2024.
\newblock URL \url{https://arxiv.org/abs/2408.06292}.

\bibitem[Yamada et~al.(2025)Yamada, Lange, Lu, Hu, Lu, Foerster, Clune, and Ha]{yamada2025aiscientistv2}
Yutaro Yamada, Robert~Tjarko Lange, Cong Lu, Shengran Hu, Chris Lu, Jakob Foerster, Jeff Clune, and David Ha.
\newblock The ai scientist-v2: Workshop-level automated scientific discovery via agentic tree search, 2025.
\newblock URL \url{https://arxiv.org/abs/2504.08066}.

\bibitem[Li et~al.(2024)Li, Xu, Guo, Zhao, Li, Yuan, Zhang, Jiang, Xin, Dang, Zhao, Rong, Feng, and Bing]{li2024chain}
Long Li, Weiwen Xu, Jiayan Guo, Ruochen Zhao, Xingxuan Li, Yuqian Yuan, Boqiang Zhang, Yuming Jiang, Yifei Xin, Ronghao Dang, Deli Zhao, Yu~Rong, Tian Feng, and Lidong Bing.
\newblock Chain of ideas: Revolutionizing research via novel idea development with llm agents, 2024.
\newblock URL \url{https://arxiv.org/abs/2410.13185}.

\bibitem[Baek et~al.(2025)Baek, Jauhar, Cucerzan, and Hwang]{baek2025researchagent}
Jinheon Baek, Sujay~Kumar Jauhar, Silviu Cucerzan, and Sung~Ju Hwang.
\newblock Researchagent: Iterative research idea generation over scientific literature with large language models, 2025.
\newblock URL \url{https://arxiv.org/abs/2404.07738}.

\bibitem[Si et~al.(2024)Si, Yang, and Hashimoto]{idea_agent}
Chenglei Si, Diyi Yang, and Tatsunori Hashimoto.
\newblock Can llms generate novel research ideas? {A} large-scale human study with 100+ {NLP} researchers.
\newblock \emph{CoRR}, abs/2409.04109, 2024.

\bibitem[Shao et~al.(2024)Shao, Jiang, Kanell, Xu, Khattab, and Lam]{storm}
Yijia Shao, Yucheng Jiang, Theodore~A. Kanell, Peter Xu, Omar Khattab, and Monica~S. Lam.
\newblock Assisting in writing wikipedia-like articles from scratch with large language models.
\newblock In \emph{{NAACL-HLT}}, pages 6252--6278. Association for Computational Linguistics, 2024.

\bibitem[Jin et~al.(2024)Jin, Zhao, Wang, Chen, Zhu, Xiao, and Wang]{jin2024agentreview}
Yiqiao Jin, Qinlin Zhao, Yiyang Wang, Hao Chen, Kaijie Zhu, Yijia Xiao, and Jindong Wang.
\newblock Agentreview: Exploring peer review dynamics with llm agents.
\newblock \emph{EMNLP}, 2024.

\bibitem[Contributors(2023)]{langchain2023}
LangChain Contributors.
\newblock Langchain: Build context-aware reasoning applications.
\newblock \url{https://github.com/langchain-ai/langchain}, 2023.

\bibitem[Shen et~al.(2023)Shen, Song, Tan, Li, Lu, and Zhuang]{shen2023HuggingGPT}
Yongliang Shen, Kaitao Song, Xu~Tan, Dongsheng Li, Weiming Lu, and Yueting Zhuang.
\newblock Hugginggpt: Solving ai tasks with chatgpt and its friends in hugging face, 2023.
\newblock URL \url{https://arxiv.org/abs/2303.17580}.

\bibitem[Xie et~al.(2023)Xie, Zhou, Cheng, Shi, Weng, Liu, Hua, Zhao, Liu, Liu, Liu, Xu, Su, Shin, Xiong, and Yu]{xie2023OpenAgents}
Tianbao Xie, Fan Zhou, Zhoujun Cheng, Peng Shi, Luoxuan Weng, Yitao Liu, Toh~Jing Hua, Junning Zhao, Qian Liu, Che Liu, Leo~Z. Liu, Yiheng Xu, Hongjin Su, Dongchan Shin, Caiming Xiong, and Tao Yu.
\newblock Openagents: An open platform for language agents in the wild, 2023.
\newblock URL \url{https://arxiv.org/abs/2310.10634}.

\bibitem[Hong et~al.(2024)Hong, Zhuge, Chen, Zheng, Cheng, Zhang, Wang, Wang, Yau, Lin, Zhou, Ran, Xiao, Wu, and Schmidhuber]{hong2024MetaGPT}
Sirui Hong, Mingchen Zhuge, Jiaqi Chen, Xiawu Zheng, Yuheng Cheng, Ceyao Zhang, Jinlin Wang, Zili Wang, Steven Ka~Shing Yau, Zijuan Lin, Liyang Zhou, Chenyu Ran, Lingfeng Xiao, Chenglin Wu, and Jürgen Schmidhuber.
\newblock Metagpt: Meta programming for a multi-agent collaborative framework, 2024.
\newblock URL \url{https://arxiv.org/abs/2308.00352}.

\bibitem[{Microsoft AutoGen Team}(2025)]{autogen2025}
{Microsoft AutoGen Team}.
\newblock Autogen: A programming framework for agentic ai.
\newblock \url{https://github.com/microsoft/autogen}, 2025.
\newblock Accessed: 2025-04-21.

\bibitem[Gao et~al.(2024)Gao, Li, Pan, Kuang, Ma, Qian, Wei, Zhang, Xie, Chen, Yao, Peng, Zhang, Zhu, Cheng, Shi, Li, Ding, and Zhou]{gao2024AgentScope}
Dawei Gao, Zitao Li, Xuchen Pan, Weirui Kuang, Zhijian Ma, Bingchen Qian, Fei Wei, Wenhao Zhang, Yuexiang Xie, Daoyuan Chen, Liuyi Yao, Hongyi Peng, Zeyu Zhang, Lin Zhu, Chen Cheng, Hongzhu Shi, Yaliang Li, Bolin Ding, and Jingren Zhou.
\newblock Agentscope: A flexible yet robust multi-agent platform, 2024.
\newblock URL \url{https://arxiv.org/abs/2402.14034}.

\bibitem[Weng et~al.(2025)Weng, Zhu, Bao, Zhang, Wang, Zhang, and Yang]{weng2025CycleResearcher}
Yixuan Weng, Minjun Zhu, Guangsheng Bao, Hongbo Zhang, Jindong Wang, Yue Zhang, and Linyi Yang.
\newblock Cycleresearcher: Improving automated research via automated review, 2025.
\newblock URL \url{https://arxiv.org/abs/2411.00816}.

\bibitem[{AI Co-Scientist Team}(2025)]{aicoscientist2025}
{AI Co-Scientist Team}.
\newblock Towards an ai co-scientist.
\newblock \url{https://storage.googleapis.com/coscientist_paper/ai_coscientist.pdf}, 2025.

\bibitem[Schmidgall et~al.(2025)Schmidgall, Su, Wang, Sun, Wu, Yu, Liu, Liu, and Barsoum]{schmidgall2025agentlaboratory}
Samuel Schmidgall, Yusheng Su, Ze~Wang, Ximeng Sun, Jialian Wu, Xiaodong Yu, Jiang Liu, Zicheng Liu, and Emad Barsoum.
\newblock Agent laboratory: Using llm agents as research assistants, 2025.
\newblock URL \url{https://arxiv.org/abs/2501.04227}.

\end{thebibliography}
\clearpage
\appendix
\section{Appendix}
In the Appendix, Section~\ref{sec:tools} provides detailed definitions of all the tools employed in the \model\ system. Sections~\ref{sec:prepare_agent} through~\ref{sec:writing_agent} elaborate on the tools and system prompt configurations utilized by the system’s components, including the \ag{Knowledge Acquisition Agent}, \ag{Resource Analyst}, \ag{Code Agent}, \ag{Advisor Agent}, and \ag{Automated Documentation Agent}. Section~\ref{sec:bench_cons} presents the detailed prompt for constructing the benchmark dataset.

\subsection{Definitions of Tools}
\label{sec:tools}
The tools utilized within the \model\ system fall into three main categories: Coding, File, and Planning. Their detailed definitions are outlined below.
\begin{longtable}{p{5.2cm}|p{1.5cm}|p{6.5cm}}
\caption{List of detailed information of tools.}
\label{tab:tool_1}\\
\toprule
\textbf{Tool Name} & \textbf{Category} & \textbf{Description}     \\
 \midrule

\texttt{gen\_code\_tree\_structure} & Coding & Generate a tree structure of the code in the specified directory. Use this function when you need to know the overview of the codebase and want to generate a tree structure of the codebase. \\
\midrule
\texttt{read\_file} & Coding & Read the contents of a file and return it as a string. Use this function when there is a need to check an existing file. \\
\midrule
\texttt{create\_directory} & Coding & Create a directory if it does not exist. Use this function when there is a need to create a new directory. \\
\midrule
\texttt{list\_files} & Coding & List all files and directories under the given path if it is a directory. Use this function when there is a need to list the contents of a directory. \\
\midrule
\texttt{run\_python} & Coding & Run a python script. \\
\midrule
\texttt{write\_file} & Coding & Write content to a file. Use this function when there is a need to write content to an existing file. \\
\midrule
\texttt{create\_file} & Coding & Create a file with the given path and content. Use this function when there is a need to create a new file with initial content. \\
\midrule
\texttt{execute\_command} & Coding & Execute a command in the system shell. Use this function when there is a need to run a system command, and execute programs. \\
\midrule
\texttt{terminal\_page\_down} & Coding & Scroll the viewport DOWN one page-length in the current terminal. Use this function when the terminal is too long and you want to scroll down to see the next content. \\
\midrule
\texttt{terminal\_page\_up} & Coding & Scroll the viewport UP one page-length in the current terminal. Use this function when the terminal is too long and you want to scroll up to see the previous content. \\
\midrule
\texttt{terminal\_page\_to} & Coding & Move the viewport to the specified page index. The index starts from 1. 

Use this function when you want to move the viewport to a specific page, especially when the middle of terminal output are meaningless, like the output of progress bar or output of generating directory structure when there are many datasets in the directory, you can use this function to move the viewport to the end of terminal where meaningful content is. \\
\midrule
\texttt{open\_local\_file} & File & Open a local file at a path in the text-based browser and return current viewport content. \\
\midrule
\texttt{page\_up\_markdown} & File & Scroll the viewport UP one page-length in the current file and return the new viewport content. \\
\midrule
\texttt{page\_down\_markdown} & File & Scroll the viewport DOWN one page-length in the current file and return the new viewport content. \\
\midrule
\texttt{find\_next} & File & Scroll the viewport to next occurrence of the search string. \\
\midrule
\texttt{find\_on\_page\_ctrl\_f} & File & Scroll the viewport to the first occurrence of the search string. This is equivalent to Ctrl+F. \\
\midrule
\texttt{question\_answer\_on\_whole\_page} & File & Ask a question on the whole page and return the answer. \\
\midrule
\texttt{visual\_question\_answering} & File & This tool is used to answer questions about attached images or videos. \\
\midrule
\texttt{plan\_dataset} & Planning & Plan the dataset for the task. Use this tool after you have carefully reviewed the existing resources and understand the task. \\
\midrule
\texttt{plan\_training} & Planning & Plan the training process for the model. Use this tool after you have carefully reviewed the existing resources and understand the task. \\
\midrule
\texttt{plan\_testing} & Planning & Plan the test process for the model. Use this tool after you have carefully reviewed the existing resources and understand the task. \\
\midrule
\texttt{plan\_testing} & Planning & Plan the test process for the model. Use this tool after you have carefully reviewed the existing resources and understand the task. \\

\bottomrule
\end{longtable}

\subsection{Knowledge Acquisition Agent}
\label{sec:prepare_agent}
The specific tools and system prompt for implementing the \ag{Knowledge Acquisition Agent} are as follows:

\begin{lstlisting}[basicstyle=\ttfamily\footnotesize, frame=none, columns=fullflexible, breaklines=true, breakatwhitespace=ture, breakindent=0pt, language=Tools, caption={Tools of \ag{Knowledge Acquisition Agent}}, frame=shadowbox,xleftmargin=0.02\linewidth, xrightmargin=0.02\linewidth]
[open_local_file, page_up_markdown, page_down_markdown, find_on_page_ctrl_f, find_next, visual_question_answering, transfer_back_to_orchestrate_agent]
\end{lstlisting}

\vspace{-0.15in}\begin{lstlisting}[basicstyle=\ttfamily\footnotesize, frame=none, columns=fullflexible, breaklines=true, breakatwhitespace=ture, breakindent=0pt, language=Prompt, postbreak=\mbox{\textcolor{gray}{$\hookrightarrow$}\space}, caption={System Prompt of \ag{Knowledge Acquisition Agent}}, frame=shadowbox,xleftmargin=0.02\linewidth, xrightmargin=0.02\linewidth]
You are given a list of papers, searching results of the papers on GitHub, and innovative ideas according to the papers. Your working directory is `/workplace`, you can only access files in this directory.

Your task is to go through the searching results, find out more detailed information about repositories in the searching results, and determine which repositories are the most relevant and useful to the innovative ideas. You can determine the relevance and usefulness by the following criteria:
1. Repositories with more stars are more recommended.
2. Repositories created more recently are more recommended, [IMPORTANT!] Too old repositories are not recommended.
3. More detaild `README.md` file means more readable codebase and more reproducible, so more recommended.
4. More clear code structure, code comments, and inline code explanations mean more readable codebase and more maintainable, so more recommended.
5. I prefer repositories with `python` language, and running coding in the local machine rather than in docker. As for deep learning projects, I prefer `pytorch` framework.

You should choose at least 5 repositories as the reference codebases.

I should use the determined repositories as reference codebases to implement the innovative ideas, so your decision should be as accurate as possible, and the number of repositories should be as less as possible. 

During the decision process, you can use the following tools:
1. You can use `execute_command` to git clone the repository to the working directory `/workplace`. Choose 5-8 repositories you really need. And you should reserve the names of the repositories.

2. You can use `gen_code_tree_structure` to generate the tree structure of the code in the repository.

3. You can use `read_file` to read the content of the file in the repository. Note that read `README.md` file can help you know the purpose and function of the code in the repository, and read other files can help you know the details of the implementation.

4. You can use `terminal_page_down`, `terminal_page_up` and `terminal_page_to` to scroll the terminal output when it is too long. You can use `terminal_page_to` to move the viewport to the specific page of terminal where the meaningful content is, for example, when the terminal output contains a progress bar or output of generating directory structure when there are many datasets in the directory, you can use `terminal_page_to` to move the viewport to the end of terminal where the meaningful content is.

4. Finally, you should use the function `case_resolved` to output the determined reference codebases.
\end{lstlisting}

\subsection{Resource Analyst}
\label{sec:resource_analyst}
The \ag{Resource Analyst} module comprises three sub-agents: the \ag{Paper Analyst} in Section~\ref{sec:paper_ana}, the \ag{Code Analyst} in Section~\ref{sec:code_ana}, and the \ag{Plan Agent} in Section~\ref{sec:plan_agent}. The \ag{Paper Analyst} and \ag{Code Analyst} extract academic concepts from research papers and their corresponding code interpretations, respectively. The \ag{Plan Agent} is responsible for generating a comprehensive development plan, encompassing dataset selection, training methodology, and evaluation procedures. The tools employed by these agents, along with their corresponding system prompts, are detailed below.

\subsubsection{Paper Analyst}
\label{sec:paper_ana}
\begin{lstlisting}[basicstyle=\ttfamily\footnotesize, frame=none, columns=fullflexible, breaklines=true, breakatwhitespace=ture, breakindent=0pt, language=Tools, caption={Tools of \ag{Paper Analyst}}, frame=shadowbox,xleftmargin=0.02\linewidth, xrightmargin=0.02\linewidth]
[open_local_file, page_up_markdown, page_down_markdown, find_on_page_ctrl_f, find_next, question_answer_on_whole_page]
\end{lstlisting}

\vspace{-0.15in}\begin{lstlisting}[basicstyle=\ttfamily\footnotesize, frame=none, columns=fullflexible, breaklines=true, breakatwhitespace=ture, breakindent=0pt, language=Prompt, postbreak=\mbox{\textcolor{gray}{$\hookrightarrow$}\space}, caption={System Prompt of \ag{Paper Analyst}}, frame=shadowbox,xleftmargin=0.02\linewidth, xrightmargin=0.02\linewidth]
You are a `Paper Survey Agent` specialized in analyzing academic papers. Your task is to extract and analyze specific academic concepts from research papers located in `/workplace/papers/`.

OBJECTIVE:
- Analyze the provided academic definition
- Extract relevant mathematical formulas and theoretical foundations
- Prepare comprehensive notes for the `Code Survey Agent`

AVAILABLE TOOLS:
1. Paper Navigation:
   - `open_local_file`: Open and read paper files
   - `page_up_markdown`/`page_down_markdown`: Navigate through pages
   - `find_on_page_ctrl_f`/`find_next`: Search specific content

2. Content Analysis:
   - `question_answer_on_whole_page`: Ask specific questions about the paper
   Example: "What is the math formula for Transformer?"

WORKFLOW:
1. Open and read the relevant papers
2. Search for the specified academic definition
3. Extract:
   - Formal definitions
   - Mathematical formulas
   - Key theoretical components
4. Document your findings and transfer your findings to the `Code Survey Agent` using the `transfer_to_code_survey_agent` function. Make sure you have read these papers thoroughly.

REQUIREMENTS:
- Be thorough in your analysis
- Focus on mathematical precision
- Ensure all extracted information is directly relevant to the given academic definition
- Provide clear and structured notes that can be effectively used by the Code Survey Agent

Remember: Your analysis forms the theoretical foundation for the subsequent code implementation phase.
\end{lstlisting}

\subsubsection{Code Analyst}
\label{sec:code_ana}

\begin{lstlisting}[basicstyle=\ttfamily\footnotesize, frame=none, columns=fullflexible, breaklines=true, breakatwhitespace=ture, breakindent=0pt, language=Tools, caption={Tools of \ag{Code Analyst}}, frame=shadowbox,xleftmargin=0.02\linewidth, xrightmargin=0.02\linewidth]
[gen_code_tree_structure, read_file, terminal_page_down, terminal_page_up, terminal_page_to]
\end{lstlisting}

\vspace{-0.15in}\begin{lstlisting}[basicstyle=\ttfamily\footnotesize, frame=none, columns=fullflexible, breaklines=true, breakatwhitespace=ture, breakindent=0pt, language=Prompt, postbreak=\mbox{\textcolor{gray}{$\hookrightarrow$}\space}, caption={System Prompt of \ag{Code Analyst}}, frame=shadowbox,xleftmargin=0.02\linewidth, xrightmargin=0.02\linewidth]
You are a `Code Survey Agent` specialized in analyzing code implementations of academic concepts. Your task is to examine codebases and match theoretical concepts with their practical implementations.

OBJECTIVE:
- Analyze codebases from reference papers in `/workplace/`
- Map academic definitions and mathematical formulas to their code implementations
- Create comprehensive implementation notes

AVAILABLE TOOLS:
1. Code Navigation:
   - `gen_code_tree_structure`: Generate repository structure overview
   - `read_file`: Access and read specific files
   - `terminal_page_down`: Scroll the viewport DOWN one page-length in the current terminal. Use this function when output of the tool is too long and you want to scroll down to see the next content.
   - `terminal_page_up`: Scroll the viewport UP one page-length in the current terminal. Use this function when output of the tool is too long and you want to scroll up to see the previous content.
   - `terminal_page_to`: Move the viewport to the specific page index. Use this function when the terminal output contains a progress bar or output of generating directory structure when there are many datasets in the directory, you can use this function to move the viewport to the end of terminal where the meaningful content is.

2. Documentation:
   - `transfer_back_to_survey_agent`: Document findings and merge with `Paper Survey Agent`'s notes

WORKFLOW:
1. Review provided academic definitions and formulas from `Paper Survey Agent`
2. Generate and analyze codebase structure
3. Locate relevant implementation files
4. Extract and document:
   - Code implementations
   - Implementation details
   - Key functions and classes
5. Merge findings with `Paper Survey Agent`'s notes and transfer complete documentation back to `Survey Agent`using the `transfer_back_to_survey_agent` function

REQUIREMENTS:
- Ensure code examples directly correspond to theoretical concepts
- Focus on critical implementation details
- Document any important variations or optimizations
- Provide clear connections between theory and implementation

Remember: Your analysis bridges the gap between theoretical concepts and practical implementation.
\end{lstlisting}

\subsubsection{Plan Agent}
\label{sec:plan_agent}
\begin{lstlisting}[basicstyle=\ttfamily\footnotesize, frame=none, columns=fullflexible, breaklines=true, breakatwhitespace=ture, breakindent=0pt, language=Tools, caption={Tools of \ag{Plan Agent}}, frame=shadowbox,xleftmargin=0.02\linewidth, xrightmargin=0.02\linewidth]
[read_file, plan_dataset, plan_training, plan_testing, gen_code_tree_structure, case_resolved, terminal_page_down, terminal_page_up, terminal_page_to]
\end{lstlisting}

\vspace{-0.15in}\begin{lstlisting}[basicstyle=\ttfamily\footnotesize, frame=none, columns=fullflexible, breaklines=true, breakatwhitespace=ture, breakindent=0pt, language=Prompt, postbreak=\mbox{\textcolor{gray}{$\hookrightarrow$}\space}, caption={System Prompt of \ag{Code Analyst}}, frame=shadowbox,xleftmargin=0.02\linewidth, xrightmargin=0.02\linewidth]
You are a Machine Learning Expert tasked with creating a detailed implementation plan for innovative ML projects.

AVAILABLE RESOURCES:
1. User's innovative idea
2. Reference codebases (in `/workplace`) selected by the `Prepare Agent`
3. Comprehensive notes from the `Survey Agent` (to be used as model plan)

WORKFLOW:
1. Code Review Phase
   - Use `gen_code_tree_structure` to understand codebase structure
   - Use `read_file` to examine specific implementations
   - Document key implementation patterns and useful components
   - Use `terminal_page_down`, `terminal_page_up` and `terminal_page_to` to scroll the terminal output when it is too long.
2. Planning Phase
   Must include these components:
   a. Dataset Plan (`plan_dataset`)
      - Dataset Description
      - Dataset Location
      - Task Definition
      - Data loading pipeline
         - Read data step
         - Data preprocessing step
         - Data dataloader step

   b. Model Plan (from Survey Agent's notes)
      - Math formula
      - Implementation details
      - Reference codebases
      - Reference papers

   c. Training Plan (`plan_training`)
      - Training pipeline
      - Loss functions
      - Optimization strategy
      - Training configurations
      - Monitoring and logging

   d. Testing Plan (`plan_testing`)
      - Test metrics
      - Test dataset preparation
      - Test code

IMPORTANT REQUIREMENTS:
1. Resource Review
   - MUST thoroughly review all provided codebases before planning
   - MUST understand the complete task scope
   - MUST analyze existing implementations for reusable components

2. Plan Generation
   - Each plan component must be detailed and actionable
   - Include specific implementation references from codebases
   - Ensure all components work together coherently

3. Testing Focus
   - Testing plan is mandatory
   - Must cover both unit tests and integration tests
   - Include specific metrics for evaluation
   - Define success criteria clearly

Your goal is to create a comprehensive, practical, and implementable plan that bridges the innovative idea with actual code implementation.
\end{lstlisting}

\subsection{Code Agent}
\label{sec:ml_agent}
The specific tools and system prompt for implementing the \ag{Code Agent} are as follows:

\begin{lstlisting}[basicstyle=\ttfamily\footnotesize, frame=none, columns=fullflexible, breaklines=true, breakatwhitespace=ture, breakindent=0pt, language=Tools, caption={Tools of \ag{Code Agent}}, frame=shadowbox,xleftmargin=0.02\linewidth, xrightmargin=0.02\linewidth]
[gen_code_tree_structure, execute_command, read_file, create_file, write_file, list_files, create_directory, run_python, case_resolved, case_not_resolved, terminal_page_down, terminal_page_up, terminal_page_to]
\end{lstlisting}

\vspace{-0.15in}\begin{lstlisting}[basicstyle=\ttfamily\footnotesize, frame=none, columns=fullflexible, breaklines=true, breakatwhitespace=ture, breakindent=0pt, language=Prompt, postbreak=\mbox{\textcolor{gray}{$\hookrightarrow$}\space}, caption={System Prompt of \ag{Code Agent}}, frame=shadowbox,xleftmargin=0.02\linewidth, xrightmargin=0.02\linewidth]
You are a machine learning engineer tasked with implementing innovative ML projects. Your workspace is: `/workplace`.

OBJECTIVE:
Create a self-contained, well-organized implementation in `/workplace/project` based on:
- The provided innovative idea
- Reference codebases (up to 5 repositories)
- The detailed implementation plan

CODE INTEGRATION PRINCIPLES:
1. Self-Contained Project
   - ALL code must reside within the project directory
   - NO direct imports from reference codebases
   - Reference code must be thoughtfully integrated into your project structure
   - Maintain consistent coding style across integrated components

2. Code Adaptation Guidelines
   - Study reference implementations thoroughly
   - Understand the core logic and algorithms
   - Rewrite and adapt code to fit your project's architecture
   - Document the origin and modifications of adapted code
   - Ensure consistent naming conventions and style

AVAILABLE TOOLS:
1. Project Structure:
   - `create_directory`: Create organized project structure
   - `create_file`, `write_file`: Write clean, documented code
   - `list_files`, `read_file`: Examine existing code
   - `terminal_page_down`, `terminal_page_up` and `terminal_page_to`: Scroll the terminal output when it is too long. You can use `terminal_page_to` to move the viewport to the specific page of terminal where the meaningful content is, for example, when the terminal output contains a progress bar or output of generating directory structure when there are many datasets in the directory, you can use `terminal_page_to` to move the viewport to the end of terminal where the meaningful content is.
2. Execution:
   - `run_python`: Run scripts without arguments
   - `execute_command`: Run with environment variables/arguments
   Note: When using `execute_command`, use `cd xx` instead of `cwd=xx`

IMPORTANT NOTES:
1. Code Integration
   - DO NOT import directly from reference codebases
   - DO adapt and integrate code thoughtfully
   - DO document code origins and modifications

2. Project Independence
   - Ensure all dependencies are explicitly declared
   - Include all necessary utility functions
   - Maintain clean separation from reference code
   - Create a truly self-contained project

3. Implementation Checklist
   - Verify each model component against the plan
   - Confirm dataset matches specifications
   - Document any deviations or modifications
   - NO shortcuts or simplifications without approval

Remember: Your goal is to create a well-organized, self-contained project that:
1. Implements EVERY component from the model plan exactly as specified
2. Uses the EXACT datasets from the plan (no toy data)
3. Thoughtfully incorporates ideas from reference implementations
4. Maintains its own coherent structure
5. You should intergrate ALL acacdemic definition and their code implementation into the project.
\end{lstlisting}

\subsection{Advisor Agent}
\label{sec:advisor_agent}
The \ag{Advisor Agent} consists of two components. The first is a multi-agent architecture composed of the \ag{Judge Agent} in Section~\ref{sec:judge_agent} and the \ag{Code Review Agent} in Section~\ref{sec:code_review_agent}, which operates after the initial implementation. The \ag{Judge Agent} is responsible for decomposing the original idea into atomic academic concepts, while the \ag{Code Review Agent} evaluates whether these atomic concepts have been correctly implemented. The second component is activated after obtaining the initial experimental results, where the \ag{Experiment Analysis Agent} in Section~\ref{sec:exp_agent} provides suggestions for code modifications and proposes directions for further experimentation. 
\subsubsection{Judge Agent}
\label{sec:judge_agent}
\begin{lstlisting}[basicstyle=\ttfamily\footnotesize, frame=none, columns=fullflexible, breaklines=true, breakatwhitespace=ture, breakindent=0pt, language=Tools, caption={Tools of \ag{Judge Agent}}, frame=shadowbox,xleftmargin=0.02\linewidth, xrightmargin=0.02\linewidth]
[transfer_to_code_review_agent]
\end{lstlisting}

\vspace{-0.15in}\begin{lstlisting}[basicstyle=\ttfamily\footnotesize, frame=none, columns=fullflexible, breaklines=true, breakatwhitespace=ture, breakindent=0pt, language=Prompt, postbreak=\mbox{\textcolor{gray}{$\hookrightarrow$}\space}, caption={System Prompt of \ag{Judge Agent}}, frame=shadowbox,xleftmargin=0.02\linewidth, xrightmargin=0.02\linewidth]
You are a advisor that can help the `Machine Learning Agent` to implement the task.

A `Machine Learning Agent` has implemented the code in the directory `/workplace/project` with the innovative ideas, but I am not sure if the implementation is correct and meets the requirements of the innovative ideas, especially some specific academic definitions.

Your job is to go through the implementation, go through the reference codebases in the directory `/workplace`, and make sure the implementation is correct and meets the requirements of the innovative ideas, especially some specific academic definitions. 

[IMPORTANT] You should carefully check whether the `Machine Learning Agent` has implemented the specific atomic idea correctly one by one based on the survey notes and the innovative idea.

After carefully checking the implementation and the reference codebases, you should use the function `case_resolved` to propose a final suggestion about the implementation.
\end{lstlisting}

\subsubsection{Code Review Agent}
\label{sec:code_review_agent}
\begin{lstlisting}[basicstyle=\ttfamily\footnotesize, frame=none, columns=fullflexible, breaklines=true, breakatwhitespace=ture, breakindent=0pt, language=Tools, caption={Tools of \ag{Code Review Agent}}, frame=shadowbox,xleftmargin=0.02\linewidth, xrightmargin=0.02\linewidth]
[read_file, gen_code_tree_structure, terminal_page_down, terminal_page_up, terminal_page_to]
\end{lstlisting}

\vspace{-0.15in}\begin{lstlisting}[basicstyle=\ttfamily\footnotesize, frame=none, columns=fullflexible, breaklines=true, breakatwhitespace=ture, breakindent=0pt, language=Prompt, postbreak=\mbox{\textcolor{gray}{$\hookrightarrow$}\space}, caption={System Prompt of \ag{Code Review Agent}}, frame=shadowbox,xleftmargin=0.02\linewidth, xrightmargin=0.02\linewidth]
You are a code reviewer, who can help me review the code in the directory: `/workplace`.

A `Machine Learning Agent` has implemented the code in the directory `/workplace/project` with the innovative ideas, and you should review the code to ensure it meets the requirements of the innovative ideas, rather than a toy implementation.

You can also review the reference codebases in the directory `/workplace` to get more information about the task.

Use `terminal_page_down` `terminal_page_up` and `terminal_page_to` to scroll the terminal output when it is too long.
[Note] You can use `terminal_page_to` to move the viewport to the end of terminal when the middle of terminal output are meaningless, like the output of progress bar or output of generating directory structure when there are many datasets in the directory, you can use this function to move the viewport to the end of terminal where the meaningful content is.

After reviewing the code, you should use the function `transfer_to_judge_agent` to transfer the conversation to the `Judge Agent`, and give a code review report.
\end{lstlisting}

\subsubsection{Experiment Analysis Agent}
\label{sec:exp_agent}
\begin{lstlisting}[basicstyle=\ttfamily\footnotesize, frame=none, columns=fullflexible, breaklines=true, breakatwhitespace=ture, breakindent=0pt, language=Tools, caption={Tools of \ag{Experiment Analysis Agent}}, frame=shadowbox,xleftmargin=0.02\linewidth, xrightmargin=0.02\linewidth]
[open_local_file, page_up_markdown, page_down_markdown, find_on_page_ctrl_f, find_next, question_answer_on_whole_page, visualizer, gen_code_tree_structure, read_file, terminal_page_down, terminal_page_up, terminal_page_to]
\end{lstlisting}

\vspace{-0.15in}\begin{lstlisting}[basicstyle=\ttfamily\footnotesize, frame=none, columns=fullflexible, breaklines=true, breakatwhitespace=ture, breakindent=0pt, language=Prompt, postbreak=\mbox{\textcolor{gray}{$\hookrightarrow$}\space}, caption={System Prompt of \ag{Experiment Analysis Agent}}, frame=shadowbox,xleftmargin=0.02\linewidth, xrightmargin=0.02\linewidth]
You are given an innovative idea and some experimental results conducted by `Machine Learning Agent` in the directory `/workspace/projects/` to implement the idea. You also have some reference codebases and papers in the working directory `/workspace`.
Your task is to: 
1. Analyze the experimental results and give a detailed analysis report about the results.
2. Analyze the reference codebases and papers, and give a further plan to let `Machine Learning Agent` to do more experiments based on the innovative idea. The further experiments could include but not limited to:
    - Modify the implementation to better fit the idea.
    - Add more experiments to prove the effectiveness and superiority of the idea. 
    - Visualize the experimental results and give a detailed analysis report about the results.
    - ANY other experiments that exsiting concurrent reference papers and codebases have done.

AVAILABLE TOOLS:
1. Project and Codebase Navigation:
    - Use `gen_code_tree_structure` to understand codebase structure
    - Use `read_file` to examine specific implementations
    - Use `terminal_page_down`, `terminal_page_up` and `terminal_page_to` to scroll the terminal output when it is too long.
2. Local file navigation:
   - `open_local_file`: Open and read paper files
   - `page_up_markdown`/`page_down_markdown`: Navigate through pages
   - `find_on_page_ctrl_f`/`find_next`: Search specific content
   - `visualizer`: use this tool to SEE the experimental results, the input should be a image or a video and a corresponding question. When the experimental results are image or video, like generated images or the visualization of the experimental results, you should use this tool to see the results and give a detailed analysis report about the results.

[IMPORTANT] You should carefully and comprehensively analyze the experimental results and the reference codebases and papers, and give a detailed analysis report about the results and the further plan by use the `case_resolved` function. DO NOT use this function before you have carefully and comprehensively analyzed the experimental results and the reference codebases and papers.
\end{lstlisting}

\subsection{Automated Documentation
Agent}
The \ag{Automated Documentation Agent} follows a three-stage workflow, exemplified by the generation of the methodology section. List~\ref{lst:structure} outlines the initial section structure; List~\ref{lst:detail} illustrates the elaboration of content guided by this structure; and List~\ref{lst:checklist} presents the review and revision process conducted according to a predefined checklist.
\label{sec:writing_agent}
\begin{lstlisting}[basicstyle=\ttfamily\footnotesize, columns=fullflexible, breaklines=true, breakatwhitespace=ture, breakindent=0pt,label={lst:structure}, caption={Prompts for generating paper section, using the methodology part as an example}, frame=shadowbox,xleftmargin=0.02\linewidth, xrightmargin=0.02\linewidth, language=Prompt]
Based on the given content, generate or revise the technical methodology structure of the proposed method, using latex format.
Current iteration: {iteration}/{self.structure_iterations}

Current structure (if exists):
{current_structure}

Content to analyze:
{content}

Guidelines for structure generation:
1. FOCUS ON TECHNICAL METHODOLOGY:
   - Include only the technical components and mechanisms of the proposed method (e.g. a machine learning model)
   - Exclude experimental settings, configurations, and evaluation procedures (which may probably occure in the content. Ignore them)

2. SECTION HIERARCHY:
   - Main section should be the name of the Proposed Method (with latex command \section{{Name_of_Proposed_Method}})
   - Use subsections for major components under the entire proposed method (e.g., encoders, architectures, learning objectives), with latex commands \subsection{{...}} and \subsubsection{{...}}
   - Use subsubsections for detailed mechanisms within major components
   - Ensure logical flow from basic components to advanced mechanisms

3. REQUIRED COMMENTS:
   Add latex comments (start with %) under the \section or \subsection or \subsubsection commands to explain the following:

   For the entire "Proposed Method" section:
   - Overview of the technical approach (what techniques are used to achieve what goal)
   - Functionalities of different components (subsections)
   - How different components (subsections) work together. The reader should get a global picture of the entire framework with this description
   
   For each subsection and subsubsection:
   - Technical purpose of this component
   - Connection to other components
   - Key technical innovations or mechanisms
   - A brief introduction to the component

   For each subsection and the entire proposed framework, give an explicit workflow chart for the specific subsection or the entire framework, using text

   For each subsection, give clear definitions on the input and output of the component, from where it get the input, and to where the output is used

4. STRUCTURE FORMAT:
   \section{{Proposed Method}}
   % [Overall method description and component relations]
   % [Input and output of the entire framework]
   % [workflow of the entire framework]
   
   \subsection{{Component 1}}
   % [Technical purpose and relations]
   % [Input and output of component 1]
   % [workflow of component 1]
   
   \subsection{{Component 2}}
   % [Technical purpose and relations]
   % [Input and output of component 2]
   % [workflow of component 2]
   
   \subsubsection{{Component 2.1}}
   % [Technical purpose and relations]

   Note that subsections are first-level modules of the proposed method. subsubsections are either 1. second-level submodules that are relatively independent and important, or 2. aspects that are important to highlight to better introduce the module.

Output only the LaTeX structure with comments as specified above. Note again that you should include only model designs using a professional writing style for academic research in AI domains, exclude any implementation details (e.g. hyperparameter configurations, coding details), experimental settings, or evaluation procedures.
\end{lstlisting}

\begin{lstlisting}[basicstyle=\ttfamily\footnotesize, columns=fullflexible, breaklines=true, breakatwhitespace=ture, breakindent=0pt, label={lst:detail},caption={Detailed section writing based on generated structures, using the methodology part as an example}, frame=shadowbox,xleftmargin=0.02\linewidth, xrightmargin=0.02\linewidth, language=Prompt]

Revise or write the following subsection of the methodology section:
\subsection{{{subsection}}}

CURRENT TEXT (if any):
{current_text}

Note: This is an iterative editing process. If current text exists:
1. Build upon and improve the existing content
2. Add missing technical details
3. Refine the writing while preserving valid technical descriptions
4. Maintain consistency with previously written parts

STRUCTURE INFORMATION:
{structure}

Note: The structure above provides high-level information about:
1. The overall architecture and components of the method
2. The purpose and role of each component
3. How components interact with each other
4. The workflow of the entire system
Use this information to understand the big picture and component relationships, NOT as writing guidelines.

NEW TECHNICAL CONTENT TO INCORPORATE:
{content}

Note: The content above contains specific technical details about:
1. Model architectures and computations
2. Mathematical formulations
3. Algorithm workflows
4. Implementation details
Use this information to write concrete technical descriptions that are missing from or can improve the current text.

REFERENCE WRITING TEMPLATE:
{writing_template}

Note: This template is for reference only. Use it to understand:
1. Common academic writing patterns (e.g., how to introduce a component, present equations, explain benefits)
2. Types of content to include (e.g., motivation, technical details, mathematical formulations)
3. Logical flow of technical presentations
4. Professional academic writing style

DO NOT:
- Follow the template word by word
- Copy its exact sentence structures
- Force your content to fit its specific format

Instead:
- Write naturally while incorporating similar elements (motivation, technical details, equations, etc.)
- Adapt the writing style to best present your specific technical content
- Maintain similar levels of technical depth and academic rigor

Requirements:
1. If current text exists:
   - Preserve valid technical content
   - Maintain consistent writing style
   - Add missing technical details
   - Improve clarity and organization
2. If starting from scratch:
   - Write comprehensive technical content
   - Follow academic writing conventions
3. In both cases:
   - Include necessary technical details from the new content
   - Ensure alignment with the structure's component descriptions
   - Use proper LaTeX formatting
   - Create smooth transitions
   - Focus on technical precision

Output the detailed LaTeX text for this subsection only.
\end{lstlisting}

\begin{lstlisting}[basicstyle=\ttfamily\footnotesize, columns=fullflexible, breaklines=true, breakatwhitespace=ture, breakindent=0pt, label={lst:checklist},caption={Review and revise the methodology section based on checklist, using the methodology part as an example}, frame=shadowbox,xleftmargin=0.02\linewidth, xrightmargin=0.02\linewidth, language=Prompt]
Review and revise the methodology section following these academic writing guidelines:

Current methodology text:
{methodology_text}

CHECKLIST FOR REVISION:

1. ACADEMIC WRITING STYLE:
   - Remove any markdown-style formatting
   - Remove any code-style documentation
   - Use formal academic language and terminology
   - Maintain consistent technical writing style throughout

2. MATHEMATICAL FORMULATION:
   - Verify correctness of all mathematical notations and equations
   - Ensure consistent variable naming
   - Check equation numbering and references
   - Avoid using too long plain text in equations

3. ACADEMIC WRITING WITH MATH:
   - Ensure that all important technical modules and mechanisms are described with math equations and well-defined math notations, even they have been well-described using natural languages
   - Avoid writing too simple math equations in non-inline equations. To address such cases, you may display 2 or 3 correlated simple equations together, or show more in-depth details for the mechanism using equations.

4. CONTENT FOCUS:
   - Reduce explanations of commonly known concepts
   - Use \cite{{}} for well-established methods instead of detailed explanations. If you don't know real papers to cite, you may also simplly describe what kind of references you are referring to.
   - Concentrate on novel contributions and key technical components
   - Ensure proper balance between overview and technical depth

5. SECTION TITLES:
   - Replace generic subsection titles with context-specific ones
   - Emphasize novelty and technical focus in titles
   - Reflect the specific application domain and unique aspects
   Examples:
   - Instead of "Embedding Layer" -> "Context-Aware Knowledge Graph Embedding"
   - Instead of "Attention Mechanism" -> "Cross-Modal Attention for Knowledge Integration"
   - Instead of "Loss Function" -> "Multi-Task Knowledge Distillation Objective"
   - But remember don't make the titles too long, just 3-6 words is fine.

Output the revised methodology section incorporating all these improvements while maintaining the core technical content. Reply your latex without any additional explanations.
\end{lstlisting}

\subsection{Prompts of Benchmark Construction}
\label{sec:bench_cons}
The development process of Scientist-Bench comprises three main stages: in Section~\ref{sec:ref_extract}, key reference papers are extracted from the target paper; in Section~\ref{sec:instruction_gen}, input instructions are constructed based on both the reference papers and the target paper; and in Section~\ref{sec:input_ano}, these input instructions undergo anonymization.

\subsubsection{Reference Extraction}
\label{sec:ref_extract}
Reference Extraction is a multi-step procedure comprising five distinct steps, accompanied by a overall task description. The corresponding prompt is presented below.

\noindent\textbf{Step 1:}
\vspace{-0.15in}\begin{lstlisting}[basicstyle=\ttfamily\footnotesize, frame=none, columns=fullflexible, breaklines=true, breakatwhitespace=ture, breakindent=0pt, language=Prompt, postbreak=\mbox{\textcolor{gray}{$\hookrightarrow$}\space}, caption={Prompt of Step 1 in \textbf{Reference Extraction}}, frame=shadowbox,xleftmargin=0.02\linewidth, xrightmargin=0.02\linewidth]
[STEP 1: Citation Pattern Analysis]
Create a statistical map of citations in the paper:
- Count citation frequency
- Track citation locations (which sections)
- Note cross-section citations
- List at least 15 most frequently cited papers

Output format:
{
    "citation_map": [
        {
            "reference": "the exact paper title in references",
            "count": number,
            "sections": ["section names"],
            "quotes": ["citation contexts"]
        }
    ]
}
\end{lstlisting}

\noindent\textbf{Step 2:}
\vspace{-0.15in}\begin{lstlisting}[basicstyle=\ttfamily\footnotesize, frame=none, columns=fullflexible, breaklines=true, breakatwhitespace=ture, breakindent=0pt, language=Prompt, postbreak=\mbox{\textcolor{gray}{$\hookrightarrow$}\space}, caption={Prompt of Step 2 in \textbf{Reference Extraction}}, frame=shadowbox,xleftmargin=0.02\linewidth, xrightmargin=0.02\linewidth]
[STEP 2: Context Analysis]
Analyze how each frequently cited paper is discussed:
- Look for influence indicators (e.g., "based on", "extends")
- Assess discussion depth
- Identify methodology-related citations

Output format:
{
    "context_analysis": [
        {
            "reference": "the exact paper title in references",
            "indicators": ["relevant phrases"],
            "depth": "detailed/moderate/brief",
            "is_method": boolean,
            "quotes": ["key contexts"]
        }
    ]
}
\end{lstlisting}

\noindent\textbf{Step 3:}
\vspace{-0.15in}\begin{lstlisting}[basicstyle=\ttfamily\footnotesize, frame=none, columns=fullflexible, breaklines=true, breakatwhitespace=ture, breakindent=0pt, language=Prompt, postbreak=\mbox{\textcolor{gray}{$\hookrightarrow$}\space}, caption={Prompt of Step 3 in \textbf{Reference Extraction}}, frame=shadowbox,xleftmargin=0.02\linewidth, xrightmargin=0.02\linewidth]
[STEP 3: Evidence Collection]
For each significant citation, identify:
- Concepts/methods borrowed
- How they were modified/improved
- Evidence of influence

Output format:
{
    "evidence": [
        {
            "reference": "the exact paper title in references",
            "borrowed": ["elements used"],
            "changes": ["modifications made"],
            "evidence": ["supporting quotes"],
            "type": "foundation/component/inspiration"
        }
    ]
}
\end{lstlisting}

\noindent\textbf{Step 4:}
\vspace{-0.15in}\begin{lstlisting}[basicstyle=\ttfamily\footnotesize, frame=none, columns=fullflexible, breaklines=true, breakatwhitespace=ture, breakindent=0pt, language=Prompt, postbreak=\mbox{\textcolor{gray}{$\hookrightarrow$}\space}, caption={Prompt of Step 4 in \textbf{Reference Extraction}}, frame=shadowbox,xleftmargin=0.02\linewidth, xrightmargin=0.02\linewidth]
[STEP 4: Impact Scoring]
Score each reference based on:
- Citation frequency (30%)
- Location importance (25%)
- Discussion depth (25%)
- Direct influence (20%)

Output format:
{
    "scores": [
        {
            "reference": "the exact paper title in references",
            "total": number,
            "breakdown": {
                "frequency": number,
                "location": number,
                "depth": number,
                "influence": number
            }
        }
    ]
}
\end{lstlisting}

\noindent\textbf{Step 5:}
\vspace{-0.15in}\begin{lstlisting}[basicstyle=\ttfamily\footnotesize, frame=none, columns=fullflexible, breaklines=true, breakatwhitespace=ture, breakindent=0pt, language=Prompt, postbreak=\mbox{\textcolor{gray}{$\hookrightarrow$}\space}, caption={Prompt of Step 5 in \textbf{Reference Extraction}}, frame=shadowbox,xleftmargin=0.02\linewidth, xrightmargin=0.02\linewidth]
[STEP 5: Final Selection]
Select and justify top 15-25 most influential papers:
- Rank based on impact scores
- Provide detailed justification
- Include specific evidence
- Explain critical importance

Output format:
{
    "top_papers": [
        {
            "reference": "the exact paper title in references",
            "rank": number,
            "type": ["methodological/component/conceptual"],
            "justification": "detailed explanation",
            "usage": "how paper was used"
        }
    ]
}
\end{lstlisting}

\noindent\textbf{Overall:}
\vspace{-0.15in}\begin{lstlisting}[basicstyle=\ttfamily\footnotesize, frame=none, columns=fullflexible, breaklines=true, breakatwhitespace=ture, breakindent=0pt, language=Prompt, postbreak=\mbox{\textcolor{gray}{$\hookrightarrow$}\space}, caption={Prompt of overall task description in \textbf{Reference Extraction}}, frame=shadowbox,xleftmargin=0.02\linewidth, xrightmargin=0.02\linewidth]
[OVERALL INSTRUCTION]
Identify the most influential references in this research paper based on three criteria:
1. Methodological Foundation - Papers that provided core methods
2. Critical Components - Papers whose specific techniques were integrated
3. Conceptual Inspiration - Papers that shaped the research direction
\end{lstlisting}
\subsubsection{Research Instruction Generation}
\label{sec:instruction_gen}
The following prompt is used to generate a level 1 input idea:
\begin{lstlisting}[basicstyle=\ttfamily\footnotesize, frame=none, columns=fullflexible, breaklines=true, breakatwhitespace=ture, breakindent=0pt, language=Prompt, postbreak=\mbox{\textcolor{gray}{$\hookrightarrow$}\space}, caption={Prompt to extract the detailed idea of a given target paper.}, frame=shadowbox,xleftmargin=0.02\linewidth, xrightmargin=0.02\linewidth]
Analyze the given research paper and write a detailed technical instruction paragraph for researchers to implement its core methodology without reading the full paper. Your instruction must include:

1. What task does the model work on
2. Core techniques/algorithms used in the paper (e.g., specific neural network architectures, optimization methods, data processing approaches)
3. Purpose and function of each major technical component
4. Implementation details for each component, such as:
   - Key parameters and configurations
   - Input/output specifications
   - Important constraints or requirements
5. Step-by-step description of how these components interact and combine into the complete system
6. Critical implementation details that affect performance

(If the examples above do not apply to the input paper, ignore the examples)

Focus only on the technical methodology and implementation aspects. Exclude background information, literature review, and experimental results. Write in a clear, sequential format that a technical researcher could follow to reproduce the core method. 

Directly write the instruction without any other words.

Don't mention the specific names of the proposed model, or exact module names that are special to this paper.
\end{lstlisting}

\subsubsection{Input Anonymization}
\label{sec:input_ano}
To mitigate the risk of the input instructions including the target paper's name—thereby potentially triggering the agent to recall information memorized during pre-training—we apply anonymization to the input instructions. The anonymization procedure comprises two components: identifying model names and eliminating potential mentions of these names within the input instructions. The corresponding prompts are presented below.
\begin{lstlisting}[basicstyle=\ttfamily\footnotesize, frame=none, columns=fullflexible, breaklines=true, breakatwhitespace=ture, breakindent=0pt, language=Prompt, postbreak=\mbox{\textcolor{gray}{$\hookrightarrow$}\space}, caption={Prompt to extract the model's name of a given target paper.}, frame=shadowbox,xleftmargin=0.02\linewidth, xrightmargin=0.02\linewidth]
Given a research paper's title and abstract, extract the name of the novel model/method introduced in the paper:

1. Look for phrases that signal a new model introduction, such as:
   - "we propose/present/introduce"
   - "our model/method/approach"
   - "called/named"
   - Model name followed by model architecture details

2. Return format:
   - If a proposed model name is found: Return only the model name
   - If you find both abbreviation and full name for the model, format them into "abbreviation, full name"
   - If no clear model name is found: Return "NO MODEL NAME FOUND"
   - You should strictly follow the requirement, and output without any other words

3. Focus only on the main proposed model:
   - Ignore baseline models
   - Ignore models from referenced papers
   - Ignore general model categories/types

Input:
- Paper Title: {paper_title}
- Paper Content: {paper_content}
\end{lstlisting}

\begin{lstlisting}[basicstyle=\ttfamily\footnotesize, frame=none, columns=fullflexible, breaklines=true, breakatwhitespace=ture, breakindent=0pt, language=Prompt, postbreak=\mbox{\textcolor{gray}{$\hookrightarrow$}\space}, caption={Remove the model name of the input instruction.}, frame=shadowbox,xleftmargin=0.02\linewidth, xrightmargin=0.02\linewidth]
Given a research paper's proposed model name and its paper title, anonymize any mentions of the model name and direct paper self-references in a paragraph.

Replace:
- Model name and variations with "the proposed model" or "the proposed approach"
- Paper self-references with "this paper" or "this study"
- Keep all other content exactly as written

Input:
- Paper Title: {paper_title}
- Model Name: {model_name}
- Paragraph: {paragraph}

Output:
- If no model name mentions found: Return "NO NEED TO PROCESS"
- If anonymization needed: Return the processed paragraph with only required replacements

Note: Only anonymize the specified model name and direct paper references. Keep all other content, including other model names and references, unchanged.
\end{lstlisting}

\end{document}